%% file: main.tex
\documentclass[11pt]{article}

\usepackage[preprint]{acl}
\usepackage{tcolorbox}
\usepackage{times}
\usepackage{latexsym}
\usepackage[T1]{fontenc}
\usepackage[utf8]{inputenc}
\usepackage{microtype}
\usepackage{hyperref}
\usepackage{url}
\usepackage{booktabs}
\usepackage{xspace}
\usepackage{graphicx}
\usepackage{textcomp}
\usepackage{multicol}
\usepackage{multirow}
\usepackage{enumitem}
\usepackage{pifont}
\usepackage{amsmath}
\usepackage{subfigure}
\usepackage{float}
\usepackage{tikz}
\usepackage{subcaption}
\usepackage{lineno}
\usepackage{makecell}
\usepackage{CJKutf8}
\usepackage{inconsolata}
\usepackage{adjustbox}
\usepackage{colortbl}
\usepackage{tabularx}
\usepackage{kotex}
\SetHangulFonts{nanummj}{nanumgt}{nanumgt}
\usepackage{fvextra}
\newcommand{\promptfile}[1]{\VerbatimInput[fontsize=\scriptsize, breaklines=true, frame=single, framesep=1.5mm]{#1}}
\fvset{breaksymbolleft={}, breaksymbolindentleft=0pt}
\usepackage{placeins}

\definecolor{darkblue}{rgb}{0, 0, 0.5}
\hypersetup{colorlinks=true, citecolor=darkblue, linkcolor=darkblue, urlcolor=darkblue}

\def
\name{\textsc{KLEAR}\xspace}

\definecolor{mycolor}{RGB}{33, 95, 154}
\definecolor{custom_red}{RGB}{228, 54, 54}

\title{KLEAR: Knowledge-Matched LLM Evaluation in Korean across \\ Application Requirements via Controlled Benchmark Transformation}

\author{
\textbf{Seoyeon Kim}$^{1}$ \quad
\textbf{Jeonghyun Kang}$^{2}$ \quad
\textbf{Hamin Koo}$^{1}$ \\
\textbf{Barom Kang}$^{2}$ \quad
\textbf{Heondeuk Lee}$^{2,\dagger}$ \quad
\textbf{Jaehyung Kim}$^{1,\dagger}$ \\[3pt]
$^{1}$Yonsei University \qquad
$^{2}$DATUMO INC. \\[3pt]
\texttt{\{yeon\_04, jaehyungk\}@yonsei.ac.kr} \\
\texttt{heondeuk.lee@selectstar.ai}
}

\begin{document}
\maketitle
\begingroup
\renewcommand{\thefootnote}{$\dagger$}
\footnotetext{Equal supervision and corresponding authorship.}
\endgroup

\input{0_abstract}
\input{1_intro}
\input{2_related}
\input{3_motivation}
\input{4_klear}
\input{5_construction}
\input{6_utility}
\input{7_conclusion}

\bibliography{custom}
\appendix
\input{8_appendix}

\end{document}

%% file: 0_abstract.tex
\begin{abstract}
Existing Korean benchmarks for Large Language Models (LLMs) evaluate performance across diverse knowledge domains, but their accuracy scores cannot distinguish whether incorrect responses arise from a lack of required knowledge or a failure to apply that knowledge correctly.
Our analysis of existing Korean benchmarks reveals that most questions can be answered through factual recall alone, while those demanding greater application also differ in required knowledge, making it difficult to isolate application ability.
To address this, we introduce \name{}, a framework that constructs \textit{knowledge-matched contrasts} by varying application demand while preserving the required knowledge.
During construction, \name{} identifies the knowledge underlying each source question as an explicit anchor bundle to guide generation of questions requiring more complex application.
Generated candidates then undergo automated and human validation, with iterative revision based on evaluation feedback. 
Experiments demonstrate that the resulting knowledge-matched sets largely satisfy the intended constraints, enabling controlled evaluation across diverse LLMs.
These evaluations reveal that increasing application demand exposes failures that persist even when the required knowledge is provided.
\end{abstract}

%% file: 1_intro.tex
\section{Introduction}

As Large Language Models (LLMs) are increasingly deployed across diverse domains and languages, reliably evaluating what they can and cannot do has become essential~\citep{chang2024survey, liang2023holistic}.
Many influential benchmarks were originally developed in English~\citep{hendrycks2020measuring}, and translation alone cannot adequately capture the knowledge, institutions, and cultural context of other languages~\citep{son2025kmmlu}.
This has motivated language-specific benchmarks built from native-language sources.
In Korean, such resources now cover diverse areas, including general language understanding~\citep{park2021klue}, cultural and linguistic knowledge~\citep{kim2024click}, and expert knowledge in academic and professional fields~\citep{son2025kmmlu, hong-etal-2025-kmmlu}.

This broader coverage alone, however, is insufficient to fully diagnose why a model fails or to indicate how it should be improved.
Results reported by subject or domain only characterize performance within each field~\citep{ribeiro2020beyond, burnell2023rethink}, but they do not reveal whether an error stems from missing knowledge or from a failure to apply knowledge the model already has. 
Moreover, because questions in existing benchmarks differ along many dimensions at once, it is difficult to attribute a performance gap to either source~\citep{gardner2020evaluating, wu2024reasoning}.

We first analyze three existing Korean benchmarks~\citep{kim2024click, hong-etal-2025-kmmlu} by labeling each question with the knowledge it requires and the application it demands.
Given that LLMs tend to underperform in non-English languages such as Korean~\citep{ahuja2023mega}, an LLM annotator is prone to misjudging the required knowledge and application when given only the question.
We therefore generate \textit{silver reasoning} by prompting a model with each question and its gold answer, following \citet{zelikman2022star}, and provide the resulting rationales as supplementary reference for annotation.
Our annotations reveal that these Korean benchmarks are predominantly composed of questions requiring knowledge recall without additional application.
Analysis of model responses further shows that these item-level labels correspond to distinct failure patterns: knowledge failures predominate on questions labeled as recall-only, whereas application failures account for a larger share of errors on questions labeled as requiring application.
However, this correspondence does not isolate the effect of application demand, as questions at different application levels also differ in knowledge content.
This motivates our central question of how to construct and validate \textit{knowledge-matched contrasts} that preserve the knowledge required by a source question while varying its application demand.

To address this question, we propose \name, a knowledge-anchored benchmark construction pipeline that holds the required knowledge fixed while systematically varying application demand.
Starting from a seed question, \name derives an \textit{anchor knowledge bundle} from its silver reasoning, which specifies the knowledge to preserve and serves as a shared basis for LLM-based generation at different target-application levels.
A shared anchor alone does not guarantee the intended comparison, because a generated question may require additional knowledge, fail to realize the target-application demand, or admit a shortcut that bypasses it.
\name therefore validates each candidate through LLM-based auditing, proxy-model difficulty screening, and targeted human validation, assessing question validity, knowledge preservation, and target-application validity. 
Feedback from these evaluations then guides iterative revision and final selection.

We evaluate \name{} in terms of construction quality and diagnostic utility.
For construction, \name{} yields 287 validated questions from 1,964 generated candidates, and removing the anchor bundle raises the share of candidates that require outside knowledge from 1.9\% to 20.4\%, showing that explicit anchoring is what keeps the required knowledge fixed.
For diagnosis, we evaluate eight LLMs on these questions with and without the required knowledge. For the three Korean LLMs on matched set, the share of application failures rises from 37.3\% to 76.9\%.
These failures persist even when the required knowledge is provided, highlighting the importance of evaluating knowledge and application separately.

Our contributions are as follows.
\begin{itemize}[leftmargin=3.5mm, topsep=2pt, itemsep=2pt, parsep=0pt]
    \item[$\circ$] We analyze the knowledge and application demands of existing Korean benchmarks and the associated model failure patterns.

    \item[$\circ$] We propose \name{}, a pipeline that constructs knowledge-matched contrasts through anchor-based generation, validation, and revision.
    
    \item[$\circ$] We show that the resulting benchmark satisfies the intended constraints and enables diagnostic comparison of model failures.
\end{itemize}

%% file: 2_related.tex
\section{Related Work}

\paragraph{Korean LLM benchmarks.}
Korean LLM evaluation has expanded from cultural and linguistic knowledge \citep{kim2024click} to expert knowledge drawn from national exams \citep{son2025kmmlu, hong-etal-2025-kmmlu}, with results reported by subject or category.
Recent Korean benchmarks have also sought to diagnose where model failures arise, through stepwise error analysis of multi-step questions \citep{son2025multi} and question-level precedents that separate legal knowledge from reasoning \citep{oh2026korean}.
Similar efforts in other domains partition existing questions into knowledge- and reasoning-focused subsets \citep{thapa2025disentangling} or decompose complex questions into sub-questions at different depths of knowledge \citep{ko2024hierarchical}.
In these approaches, however, questions with different reasoning or application demands generally also require different knowledge. 
We instead analyze existing Korean benchmarks under a shared knowledge--application framework and construct contrasts in which the required knowledge is fixed.

\paragraph{Controlled and LLM-based benchmark construction.}
Controlled evaluation changes one factor while keeping the rest of a question intact, through minimal edits \citep{gardner2020evaluating}, counterfactual conditions \citep{wu2024reasoning}, or templated variants \citep{mirzadeh2025gsm}.
LLM-based pipelines scale construction by generating questions from declared desiderata \citep{li2025autobencher} or taxonomies \citep{davidson2026reasoning}.
Closest to our setting, some methods extend existing benchmarks from seed questions, across cognitive levels \citep{cao2024structeval} or reasoning complexity \citep{zhang2024darg}, but their validation focuses mainly on answer correctness.
In contrast, \name{} holds the required knowledge constant and manipulates application demand, and verifies knowledge preservation and target-application validity through LLM-based and human validation. 

%% file: 3_motivation.tex
\section{Motivation}
\label{sec:motivation}

We first examine what existing Korean benchmarks demand of a model and where models fail on them.
We analyze 7,340 multiple-choice questions from CLIcK~\citep{kim2024click}, KMMLU-Redux, and KMMLU-Pro~\citep{hong-etal-2025-kmmlu}.

\subsection{Silver Reasoning}
\label{sec:silver}

Identifying the knowledge a question requires and how that knowledge must be applied is challenging for an LLM annotator given only the question, since LLMs tend to underperform in non-English languages such as Korean \citep{koo2026emcee}.
To support this assessment, we follow the answer-conditioned rationalization approach of \citet{zelikman2022star} and use GPT-5.4~\citep{openai2026gpt54} to generate step-by-step solutions conditioned on the question, options, and gold answer, including explanations for incorrect options. 
We refer to these model-generated solutions as \textit{silver reasoning}, as they have not been verified by domain experts.

We assess answer recoverability on 150 sampled solutions (50 per benchmark; Appendix~\ref{app:silver}).
After removing explicit statements of the final answer from the silver reasoning, we ask a verifier (GPT-5.4) to answer each question given its options and silver reasoning. 
The verifier answers all 150 questions correctly, compared with 118 when given only the question and options. 
While this suggests that silver reasoning helps the verifier answer the questions correctly, it does not establish the factual correctness of individual reasoning steps.
We therefore use silver reasoning as supplementary reference material for the annotations below and as a source for deriving anchor knowledge in Sec.~\ref{sec:klear}, rather than as ground-truth solutions. 

\subsection{Knowledge and Application Demands}
\label{sec:ka}
\input{Tables/Table_ka_val}

We characterize each question along two axes.
\begin{itemize}[leftmargin=3.5mm, topsep=2pt, itemsep=2pt, parsep=0pt]
    \item[$\circ$] \textit{Knowledge demand} (K) is the level of prior knowledge needed to solve the question: none (K0), everyday common sense (K1), school-level knowledge (K2), or specialized professional knowledge (K3).
    \item[$\circ$] \textit{Application demand} (A) is the complexity of connecting that knowledge to the case described in the question. 
    At A0, the retrieved fact or rule is itself the answer. 
    At A1, the correspondence between the knowledge and the case is stated explicitly.
    At A2, a single piece of knowledge applies, but one must judge whether the case satisfies its condition.
    At A3, two or more candidate rules compete, and one must decide which governs the case.
\end{itemize}
Both axes describe the question and its solution, independent of how the model performs.
We use GPT-5.4 to assign K and A, given the question, options, gold answer, and silver reasoning.
See Tables~\ref{tab:k_examples} and~\ref{tab:a_examples} for examples.

To validate these labels, two human annotators independently labeled 60 questions sampled evenly across the four A levels, seeing only the question, options, and gold answer (Appendix~\ref{app:ka_annotation}).
Table~\ref{tab:ka_val} compares their labels with those of the model, with and without silver reasoning.
With silver reasoning, the model agrees with the annotators on A ($\kappa=0.69$) more closely than the annotators agree with each other ($\kappa=0.60$).
Without silver reasoning, agreement on A drops to 0.47. 
Silver reasoning does not improve agreement on K, where the model tends to assign K2 to questions that annotators consider K3. 
Since application demand is the axis our analysis and construction focus on, we adopt silver reasoning for annotation.

\input{Tables/Table_ka_dist}

Table~\ref{tab:ka_dist} shows the resulting distribution.
Most questions demand specialized knowledge but little application:
61.7\% are K3 and 61.1\% are A0, and K3 questions answerable by recall alone account for 39.6\% of the total.
A0 is the largest level in every benchmark, ranging from 50.6\% in KMMLU-Pro to 73.6\% in KMMLU-Redux.
Questions that require application are fewer and unevenly distributed, with 72.8\% of A3 questions coming from KMMLU-Pro.

\input{Figures/Figure1}

\subsection{Failure Analysis}
\label{sec:failure}

We next examine how models fail on these questions.
We evaluate three Korean LLMs, K-EXAONE \citep{choi2026k}, A.X K1 \citep{cheon2026ax}, and Solar Open \citep{park2026solar}, with reasoning enabled, and collect their 5,808 incorrect responses.
For each response, GPT-5.4 reads the model's reasoning trace alongside the silver reasoning and assigns one or more failure types, ordered by their contribution to the error (Appendix~\ref{app:source_failure}).
We group these types into two categories.
\begin{itemize}[leftmargin=3.5mm, topsep=2pt, itemsep=2pt, parsep=0pt]
    \item[$\circ$] \textit{Knowledge failure}: the model recalls an incorrect fact, attributes a fact to the wrong entity, or never engages the required rule.
    \item[$\circ$] \textit{Application failure}: the model invokes the relevant knowledge but misapplies it, for example by misjudging a condition or an exception.
\end{itemize}
The remaining responses stem from invalid outputs or defective questions.
We classify each response by its primary failure type. 

\input{Figures/Figure2}

Figure~\ref{fig:failure} shows that failure composition differs between questions answerable by recall alone and questions that require application.
On A0 questions, knowledge failures dominate (69.8\% vs.\ 21.1\% application failures).
On A1--A3 questions, application failures account for 36.6--44.0\% of incorrect responses, comparable to knowledge failures (40.5--44.8\%).
The same pattern holds among K3 questions alone (Appendix~\ref{app:source_failure}).
Among questions that require application, accuracy also falls from 88.5\% at A1 to 60.0\% at A3.

These comparisons, however, are not controlled, because questions at different A levels also differ in the knowledge they require.
For instance, the share of K3 questions ranges from 40.3\% at A1 to 77.3\% at A3, and a single benchmark of professional licensure exams accounts for 72.8\% of A3 questions but at most 33.4\% at any other level. 
A gap between A levels may thus reflect what the questions are about rather than how much application they demand.
Isolating the effect of application demand requires contrasts in which the required knowledge is held fixed, which we construct in Sec.~\ref{sec:klear}.

%% file: Tables/Table_ka_val.tex
\begin{table}[t]
    \centering
    \small
    \caption{\textbf{Validation of knowledge (K) and application (A) labels.} LLM--human agreement on 60 questions, averaged across two human annotators: exact match (EM; \%) and linearly weighted Cohen's $\kappa$, which penalizes disagreements in proportion to the distance between levels. \textit{w/o silver} and \textit{w/ silver} denote the LLM annotation without and with silver reasoning.}
    \label{tab:ka_val}

    \begin{tabular}{l|cc|cc}
        \toprule[1.5pt]
        & \multicolumn{2}{c|}{\textbf{K}} & \multicolumn{2}{c}{\textbf{A}} \\
        \cmidrule(lr){2-3} \cmidrule(lr){4-5}
        & EM & $\kappa$ & EM & $\kappa$ \\
        \midrule[1pt]
        w/o silver & 68.3 & 0.67 & 45.0 & 0.47 \\
        \rowcolor[HTML]{D9D9D9}
        w/ silver & 56.7 & 0.56 & 69.2 & 0.69 \\
        \bottomrule[1.5pt]
    \end{tabular}
    \vspace{-0.1in}
\end{table}

%% file: Tables/Table_ka_dist.tex
\begin{table}[t]
    \centering
    \caption{\textbf{Distribution of knowledge (K) and application (A) demands.} Percentage of the 7,340 questions across K and A levels in three Korean benchmarks.}
    \label{tab:ka_dist}

    \resizebox{0.8\columnwidth}{!}{%
        \begin{tabular}{l|cccc|c}
            \toprule[1.5pt]
            & \textbf{A0} & \textbf{A1} & \textbf{A2} & \textbf{A3} & \textbf{Total} \\
            \midrule[1pt]
            \textbf{K0} & 0.0  & 1.0  & 0.3  & 0.0  & 1.3  \\
            \textbf{K1} & 2.9  & 1.2  & 1.1  & 0.1  & 5.2  \\
            \textbf{K2} & 18.5 & 4.0  & 5.5  & 3.6  & 31.7 \\
            \textbf{K3} & 39.6 & 4.2  & 5.1  & 12.7 & 61.7 \\
            \midrule
            \rowcolor[HTML]{D9D9D9}
            \textbf{Total} & 61.1 & 10.4 & 12.0 & 16.4 & 100.0 \\
            \bottomrule[1.5pt]
        \end{tabular}%
    }
    \vspace{-0.1in}
\end{table}

%% file: Figures/Figure1.tex
\begin{figure*}[t]
\centering
\includegraphics[width=\textwidth]{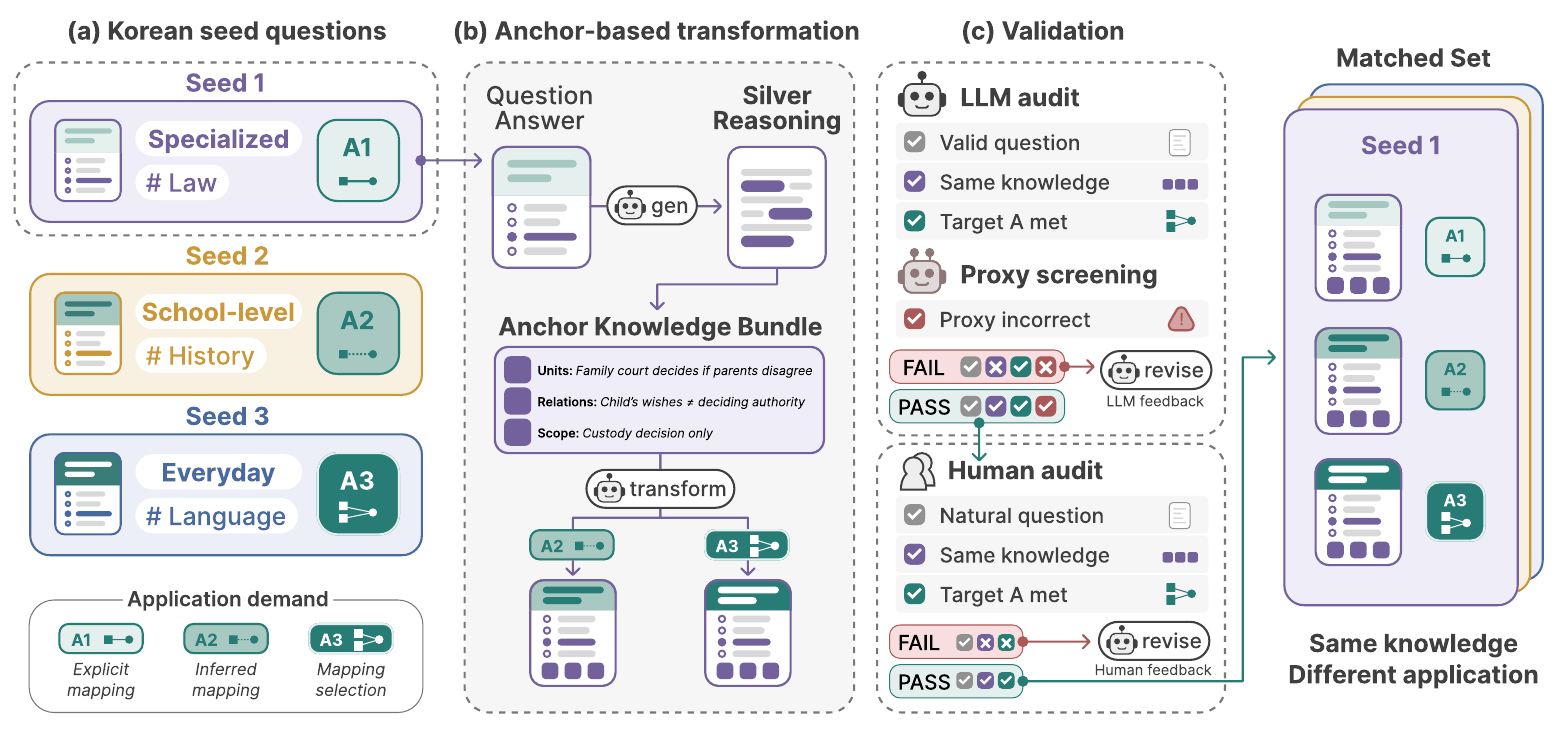}
%\vspace{-0.2in}
\caption{
\textbf{Overview of \name{}.}
\textbf{(a)} Seed questions from Korean benchmarks are labeled by knowledge (K) and application (A) demands.
\textbf{(b)} Silver reasoning is generated from the seed question and gold answer, then used to derive an anchor knowledge bundle for generating higher-A questions.
\textbf{(c)} Generated questions undergo LLM auditing, proxy screening, and human validation, with failed candidates revised using LLM or human feedback.
The result is a matched set of questions that share the same knowledge but differ in application demand.
}\label{fig:overview}
\vspace{-0.1in}
\end{figure*}

%% file: Figures/Figure2.tex
\begin{figure}[t]
\centering
\includegraphics[width=\columnwidth]{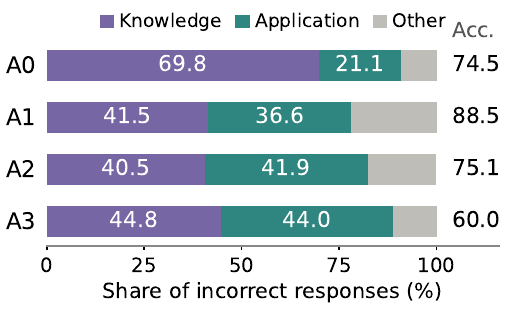}
\vspace{-0.2in}
\caption{
\textbf{Failure composition by application demand.}
Distribution of incorrect responses by failure type across three Korean LLMs, with accuracy (\%) on the right.
}\label{fig:failure}
\vspace{-0.15in}
\end{figure}

%% file: 4_klear.tex
\section{\name{}}
\label{sec:klear}

In this section, we propose \textbf{\name{}} (\textbf{K}nowledge-Matched \textbf{L}LM \textbf{E}valuation in Korean across \textbf{A}pplication \textbf{R}equirements), a framework that constructs knowledge-matched contrasts from existing benchmark questions in three stages (Figure~\ref{fig:overview}).
First, anchor-based generation (Sec.~\ref{sec:generation}) makes the knowledge behind a source question explicit as an \textit{anchor bundle}, and generates new questions that rest on the same bundle while demanding more application.
Second, automated audit and revision (Sec.~\ref{sec:audit}) uses rubric-based LLM evaluation to check the validity and quality of each candidate, screens its difficulty with a proxy model, and revises failing candidates.
Third, human validation and revision (Sec.~\ref{sec:human}) makes a final check of its quality, and revises rejected questions with annotator rationales.
By fixing the knowledge before generation and checking its preservation afterwards, KLEAR isolates differences in application demand between source questions and their contrasts.

\subsection{Anchor-Based Generation}
\label{sec:generation}

\paragraph{Seed selection.}
We draw seeds from the questions profiled in Sec.~\ref{sec:motivation}.
We select A1 or A2 questions that at least two of the three Korean LLMs answer incorrectly, so that construction starts from questions these models find difficult.
An A1 seed can target A2 or A3, whereas an A2 seed can target A3.

\paragraph{Anchor bundle.}
For each seed, an LLM constructs an \textit{anchor bundle} using the question, answer options, gold answer, and silver reasoning.
The bundle consists of three components: 
(1) \textit{Knowledge units} specify the rules, criteria, definitions, or facts required to solve the question, including those underlying the distractors.
(2) \textit{Relations} describe how these units interact, such as which unit is the principle and which the exception.
(3) A \textit{Scope note} explicitly identifies knowledge excluded from the bundle.
Together, these components define a closed set of domain knowledge that subsequent generation must stay within.
The LLM also assesses whether the bundle can support each target-application level, excluding unsupported targets from generation.
For example, an A3 target requires at least two knowledge units that could plausibly govern the same case.
% We call a generated question \textit{knowledge-matched} to its seed when both can be answered from the same anchor bundle and the case alone.

\paragraph{Target-A generation.}
Given a seed, its anchor bundle, and a target-application level, the generator LLM  produces a new multiple-choice question with four or five answer options.
The generated question must preserve the answer type of the seed and must not state or paraphrase the anchor knowledge.
For an A2 target, exactly one knowledge unit from the bundle applies, but its applicability must be inferred from the case facts.
For an A3 target, two or more knowledge units from the bundle compete over a single case with comparable supporting evidence.
Each distractor must represent an identifiable error and all options must be parallel in form.
We generate three candidates per seed--target pair.

\subsection{Automated Audit and Revision}
\label{sec:audit}

\paragraph{LLM rubric.}
An LLM auditor receives the seed, its anchor bundle, the target-application level, and the candidate.
It evaluates the candidate against eleven criteria, labeling each as \textsc{pass}, \textsc{fail}, or \textsc{uncertain} with a rationale.
L1--L6 assess question validity, while L7--L11 assess knowledge preservation and other quality criteria.
The full rubric is provided in Table~\ref{tab:rubrics}.

\paragraph{Proxy screening.}
The rubric does not measure difficulty, and a candidate can pass every criterion while being easy to answer.
We therefore use a weaker proxy model to screen for insufficiently challenging candidates.
This proxy answers each candidate from the question and answer options alone, and a candidate it answers correctly is treated as too easy.

\paragraph{Self-revision.}
A candidate is returned to the generator when any criterion is not \textsc{pass} or when the proxy answers it correctly.
If any criterion is not \textsc{pass}, the generator receives every such criterion with its rationale and must resolve them without regressing other criteria.
If all criteria are \textsc{pass} but the proxy answers correctly, the generator is told that a weaker model answered the question correctly and must close the shortcut.
In both cases, the revision must stay within the bundle.
Each candidate is revised at most once, and the revision is audited and screened again under the same conditions.
A candidate passes the automated stage when, in its original or revised form, all eleven criteria are \textsc{pass} and the proxy answers it incorrectly.

\subsection{Human Validation and Revision}
\label{sec:human}

\paragraph{Targeted validation.}
To identify defects that may remain after automated auditing, human annotators evaluate each candidate that passes the automated stage against five criteria in two stages (Table~\ref{tab:rubrics}).
In the first stage, without the anchor bundle, H1--H3 check the surface quality of the question, and in the second stage, with the bundle revealed, H4--H5 check the control of knowledge and application.
Annotators label each criterion as \textsc{pass}, \textsc{borderline}, or \textsc{fail}, with a rationale for any label other than \textsc{pass}.
A candidate passes human validation when every annotator labels all five criteria as \textsc{pass}.

\paragraph{Human-guided revision.}
Candidates that fail to pass are returned to the generator with the human annotators' labels and rationales.
The generator revises each candidate to address the identified defects while remaining within the anchor bundle and preserving the target-application level.
The revised candidates undergo human validation again under the same two-stage protocol.
Only candidates that pass human validation, either initially or after revision, are retained in the final benchmark.

%% file: 5_construction.tex
\section{Construction Quality}
\label{sec:construction_quality}
\input{Tables/Table_stats}
\input{Tables/Table_yield}
Next, we examine whether \name{} produces valid knowledge-matched contrasts and how much each validation and revision stage contributes.

\subsection{Experimental Setup}
\label{sec:construction_setup}

We use 243 seeds, 78 at A1 and 165 at A2, which form 295 seed--target pairs: 78 A1$\rightarrow$A2, 52 A1$\rightarrow$A3, and 165 A2$\rightarrow$A3.
We use GPT-5.5 to construct anchor bundles, generate and revise candidates, and audit them, and GPT-5-nano as the proxy model.
For each seed--target pair, we first generate three candidates and, if fewer than two pass auditing and screening, generate six more in one additional round.
We additionally examine how seed difficulty affects the results in Appendix~\ref{app:easy}.

Human validation proceeds in two batches.
In the first batch (164 pairs), up to two passing candidates per pair are validated.
Two annotators judge all five criteria, and two additional annotators double-check H4 and H5.
In the second batch (131 pairs), 111 pairs have a passing candidate; because of annotation budget and time constraints, we validate one candidate for 70 of them, each judged by one of three annotators. 
All annotators are native Korean speakers who work as professional data annotators and follow written guidelines calibrated on a pilot set.
Table~\ref{tab:stats} summarizes the final dataset of 287 questions built from 165 source questions.

\subsection{Construction Validity and Refinement}
\label{sec:construction_results}

\paragraph{Yield and validity.}
We first examine how many candidates survive each stage and whether the surviving ones are valid contrasts (Table~\ref{tab:yield}).
\name{} generates 1,964 candidates for the 295 pairs, of which 623 (31.7\%) pass the automated stage.
At least one candidate passes for 243 pairs (82.4\%), and human validation covers 202 of them with 301 candidates.
Of these candidates, 252 (83.7\%) pass human validation directly, and 35 more pass after human-guided revision, giving 287 final questions.
The two criteria that define a knowledge-matched contrast hold for nearly all validated candidates: knowledge sufficiency (H4) for 98.0\% and target-level structure (H5) for 95.7\%.
Although only a minority of candidates survive the automated stage, those that do largely satisfy the intended controls.

\input{Tables/Table_matched}

\paragraph{Effect of the anchor bundle.}
We examine what the anchor bundle contributes by comparing candidates generated with and without it for the same pairs.
In both cases, the generator is instructed to rely only on the knowledge of the seed, but without the bundle, 20.4\% of candidates require domain knowledge beyond it (L4), compared with 1.9\% with the bundle.
These candidates introduce knowledge that the seed never required, such as a rule that the bundle's scope note explicitly excludes.
An instruction to preserve knowledge is thus not enough, and the knowledge must be specified explicitly.

\paragraph{Automated defects and self-revision.}
Turning to the validation stages, we examine where candidates fail in the automated stage and how much self-revision helps (Table~\ref{tab:criteria}).
Before revision, only 15.0\% of candidates pass the automated stage.
L1--L5 each fail for at most 2.2\% of candidates, whereas L6--L11 each fail for 6.6--33.5\%. 
Self-revision more than doubles the number of passing candidates, from 294 to 623, although only 19.8\% of the 1,662 revised candidates pass.
It resolves 64--100\% of the failures on each criterion but also introduces new ones, most often on shortcut resistance (L8) and anchor leakage (L9).
The main bottleneck is thus not basic validity but imposing the intended knowledge and application demands.

\paragraph{Residual defects and human-guided revision.}
Finally, we examine the defects that remain after the automated stage.
Although all 301 validated candidates satisfy every LLM criterion, annotators find at least one defect in 49 (16.3\%).
These defects are mostly isolated, with 37 of the 49 failing a single criterion, and readability (H1) is the most frequent.
Human-guided revision recovers 35 of the 49 candidates (71.4\%) and resolves 77--100\% of the failures on each criterion, including all six on knowledge sufficiency (H4).
Of the 14 candidates that are discarded, 8 fail on a criterion that the revision newly introduced.
Automated auditing alone is therefore not sufficient, and revised questions need to be validated again.

%% file: Tables/Table_stats.tex
\begin{table}[t]
\centering
\small
\caption{\textbf{Final dataset statistics.} Number of source questions and generated questions by transformation.}
\label{tab:stats}
\begin{tabular}{@{}lrrrr@{}}
\toprule
 & & \multicolumn{2}{c}{\textbf{KMMLU}} & \\
\cmidrule(lr){3-4}
 & \textbf{CLIcK} & \textbf{Pro} & \textbf{Redux} & \textbf{Total} \\
\midrule
Source questions    & 66  & 53 & 46 & 165 \\
\midrule
A1$\rightarrow$A2   & 29  & 20 & 29 & 78  \\
A1$\rightarrow$A3   & 24  & 15 & 12 & 51  \\
A2$\rightarrow$A3   & 77  & 55 & 26 & 158 \\
\midrule
Generated questions & 130 & 90 & 67 & 287 \\
\bottomrule
\end{tabular}
\end{table}

%% file: Tables/Table_yield.tex
\begin{table}[t]
\centering
\small
\caption{\textbf{Stage-wise yield.}
Indented rates are relative to the preceding unindented count; the overall automated-stage pass rate is relative to all generated candidates.}
\label{tab:yield}
\begin{tabular}{@{}lrr@{}}
\toprule
\textbf{Stage} & \textbf{Count} & \textbf{Rate} \\
\midrule
\multicolumn{3}{@{}l}{\textit{Automated stage}} \\
\cmidrule{1-3}
Candidates generated                    & 1,964 & --     \\
\quad All eleven criteria \textsc{pass} & 627   & 31.9\% \\
\quad Proxy answers incorrectly         & 760   & 38.7\% \\
\quad Both (pass before revision)       & 294   & 15.0\% \\
\cmidrule{1-3}
Candidates revised                      & 1,662 & --     \\
\quad Pass after self-revision          & 329   & 19.8\% \\
\cmidrule{1-3}
Pass the automated stage                & 623   & 31.7\% \\
\midrule[\heavyrulewidth]
\multicolumn{3}{@{}l}{\textit{Human validation}} \\
\cmidrule{1-3}
Candidates validated                    & 301   & --     \\
\quad Pass directly                     & 252   & 83.7\% \\
\cmidrule{1-3}
Candidates revised                      & 49    & --     \\
\quad Pass after human-guided revision  & 35    & 71.4\% \\
\cmidrule{1-3}
Final questions                         & 287   & --     \\
\bottomrule
\end{tabular}
\vspace{-0.1in}
\end{table}

%% file: Tables/Table_matched.tex
\begin{table*}[t]
\centering
\small
\caption{\textbf{Closed-book accuracy (\%) by application level.}
\textit{Matched} covers the 30 seeds that have both an A2 and an A3 contrast (40 and 43 questions), averaged over seeds.
The numbers of A2 and A3 questions are 29 and 101 for CLIcK, 20 and 70 for KMMLU-Pro, and 29 and 38 for KMMLU-Redux.}
\label{tab:matched}
\begin{tabular}{@{}l|rrrrrrrr|rr@{}}
\toprule
 & & & \multicolumn{4}{c}{\textbf{KMMLU}} & & & & \\
\cmidrule(lr){4-7}
 & \multicolumn{2}{c}{\textbf{CLIcK}} & \multicolumn{2}{c}{\textbf{Pro}} & \multicolumn{2}{c}{\textbf{Redux}} & \multicolumn{2}{c}{\textbf{All}} & \multicolumn{2}{c}{\textbf{Matched}} \\
\cmidrule(lr){2-3} \cmidrule(lr){4-5} \cmidrule(lr){6-7} \cmidrule(lr){8-9} \cmidrule(l){10-11}
\textbf{Model} & \textbf{A2} & \textbf{A3} & \textbf{A2} & \textbf{A3} & \textbf{A2} & \textbf{A3} & \textbf{A2} & \textbf{A3} & \textbf{A2} & \textbf{A3} \\
\midrule
K-EXAONE          & 65.5  & 46.5 & 50.0  & 37.1 & 34.5 & 57.9 & 50.0 & 45.5 & 46.7 & 40.0 \\
A.X K1            & 72.4  & 40.6 & 45.0  & 27.1 & 48.3 & 47.4 & 56.4 & 37.3 & 51.7 & 21.7 \\
Solar Open        & 55.2  & 29.7 & 30.0  & 21.4 & 34.5 & 36.8 & 41.0 & 28.2 & 38.3 & 23.3 \\
\midrule
Aya Expanse 32B   & 31.0  & 26.7 & 25.0  & 21.4 & 20.7 & 18.4 & 25.6 & 23.4 & 33.3 & 20.0 \\
\midrule
GLM-4.7           & 75.9  & 67.3 & 55.0  & 48.6 & 58.6 & 63.2 & 64.1 & 60.3 & 65.0 & 56.7 \\
Qwen3.5           & 82.8  & 72.3 & 75.0  & 68.6 & 72.4 & 60.5 & 76.9 & 68.9 & 76.7 & 58.3 \\
\midrule
Claude Sonnet 5.5 & 100.0 & 86.1 & 100.0 & 77.1 & 79.3 & 78.9 & 92.3 & 81.8 & 93.3 & 81.7 \\
GPT-6.1 Sol       & 96.6  & 94.1 & 100.0 & 92.9 & 79.3 & 86.8 & 91.0 & 92.3 & 93.3 & 93.3 \\
\bottomrule
\end{tabular}
\vspace{-0.1in}
\end{table*}

%% file: 6_utility.tex
\section{Diagnostic Utility}
\label{sec:diagnostic}
\input{Figures/Figure3}
\input{Figures/Figure4}

Finally, we examine whether the contrasts built by \name{} can separate models' failures to recall knowledge from their failures to apply it.

\subsection{Experimental Setup}
\label{sec:diagnostic_setup}

We evaluate eight models\footnote{Reasoning is enabled for all models except Aya Expanse 32B.} on the 287 final questions from Sec.~\ref{sec:construction_quality}, covering four groups.
\begin{itemize}[leftmargin=3.5mm, topsep=2pt, itemsep=2pt, parsep=0pt]
    \item[$\circ$] \textit{Korean-specific}: K-EXAONE~\citep{choi2026k}, A.X K1~\citep{cheon2026ax}, and Solar Open~\citep{park2026solar}, as in Sec.~\ref{sec:failure}.
    \item[$\circ$] \textit{Multilingual-focused}: Aya Expanse 32B~\citep{dang2024aya}.
    \item[$\circ$] \textit{Open-weight frontier}: GLM-4.7~\citep{zai2025glm47} and Qwen3.5-397B-A17B~\citep{qwen2026qwen35}.
    \item[$\circ$] \textit{Proprietary frontier}: Claude Sonnet 5.5~\citep{anthropic2026sonnet55} and GPT-6.1 Sol~\citep{openai2026gpt61sol}.
\end{itemize}

Each question is answered under three conditions: \textit{closed-book}, with only the question and answer options; \textit{knowledge} (+K), with three facts drawn from the knowledge units of the anchor bundle; and \textit{knowledge and application} (+K+A), with further statements drawn from the relations of the bundle.
Both additions are written without access to the question and are shared by all questions built from the same seed.

For matched comparisons, we use the 30 seeds that have both an A2 and an A3 contrast.
All of the seeds are A1, so together with the seed question they form a complete A1--A2--A3 ladder.
%We label the incorrect closed-book responses of the three Korean LLMs with the procedure of Sec.~\ref{sec:failure}, using the solution written during construction in place of silver reasoning.

\subsection{Performance across Application Levels}
\label{sec:diagnostic_performance}

Table~\ref{tab:matched} reports closed-book accuracy across application levels.
Across all questions, seven of the eight models achieve lower accuracy on A3 than on A2.
Since most questions at the two levels come from different seeds, we also compare them on the 30 seeds that have both an A2 and an A3 contrast.
On the matched A1--A2--A3 ladders, where the underlying knowledge is held fixed, all eight models perform no better on A3 than on A2.
This consistent pattern suggests that increasing application demand makes questions more challenging even when the required knowledge is controlled.

\subsection{Failure Diagnosis and Intervention}
\label{sec:diagnostic_failure}

\paragraph{Failure profiles.}
On the matched A1--A2--A3 ladders, the share of application failures of the three Korean LLMs rises from 37.3\% at A1 to 60.0\% at A2 and 76.9\% at A3, while the share of knowledge failures falls from 41.8\% to 14.3\% (Figure~\ref{fig:ladder}).
The same shift appears across all 165 seeds, where application failures account for 32.5--46.6\% of incorrect responses on the seeds and 67.2--78.8\% on their contrasts.
This suggests that when the required knowledge is controlled, increasing application demand shifts failures from knowledge to application.

\paragraph{Knowledge and application intervention.}
Figure~\ref{fig:intervention} shows how accuracy changes as knowledge and application statements are provided.
Providing knowledge resolves most errors at A2, where accuracy reaches 83.3--100\% for all models except Aya Expanse 32B, but leaves more errors unresolved at A3, where accuracy remains at 62.7--94.3\%.
Adding the application statements then improves A3 accuracy for all eight models, by 2.9 to 15.3 points, with smaller gains at A2.
These results suggest that knowledge alone does not suffice when application demand is high, and that many of the remaining errors lie in applying the knowledge rather than in lacking it.

%% file: Figures/Figure3.tex
\begin{figure}[t]
\centering
\includegraphics[width=0.8\columnwidth]{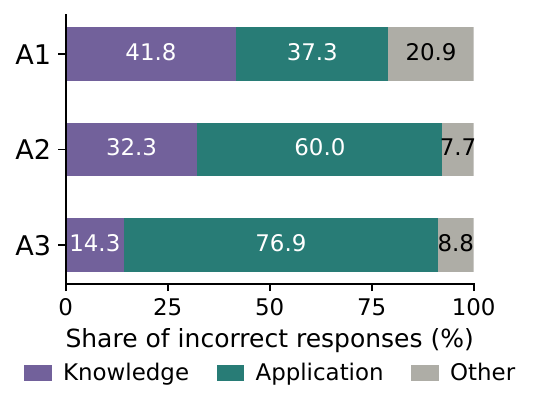}
\vspace{-0.15in}
\caption{
\textbf{Failure composition on the A1--A2--A3 ladder.}
Incorrect closed-book responses by failure type across three Korean LLMs.
A1 corresponds to the seed. \textit{Other} includes invalid outputs and defective questions.
}
\label{fig:ladder}
\vspace{-0.15in}
\end{figure}

%% file: Figures/Figure4.tex
\begin{figure*}[t]
\centering
\includegraphics[width=\textwidth]{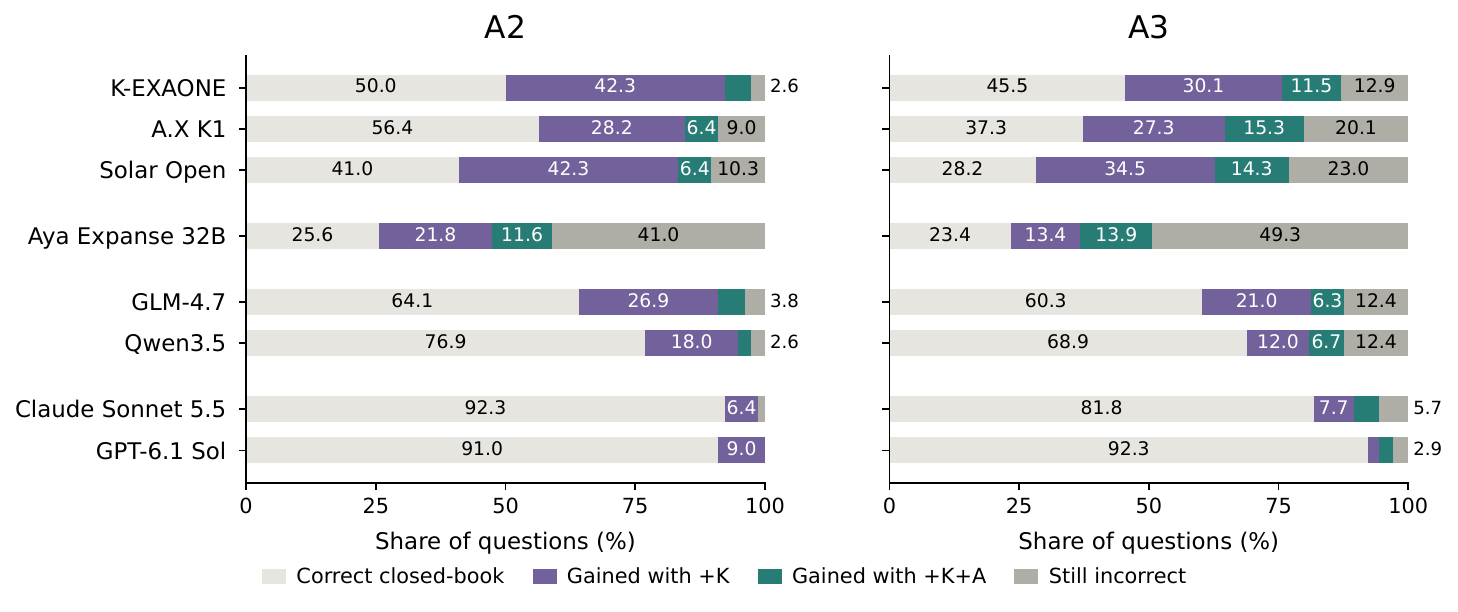}
\vspace{-0.25in}
\caption{
\textbf{Effects of knowledge and application interventions.}
Accuracy changes across closed-book, knowledge (+K), and knowledge and application (+K+A) conditions for A2 (78 questions) and A3 (209 questions).
Each bar shows closed-book accuracy, the additional gains from +K and +K+A, and the remaining error rate.
}
\label{fig:intervention}
\vspace{-0.1in}
\end{figure*}

%% file: 7_conclusion.tex
\section{Conclusion}
We propose \name{}, a framework for disentangling knowledge and application abilities in LLM evaluation through knowledge-matched contrasts.
Our analysis of existing Korean benchmarks reveals a predominance of recall-only questions and the difficulty of isolating application demand.
Through anchor-based generation and iterative validation, \name{} constructs controlled contrasts that vary application demand while preserving required knowledge. 
Evaluations across diverse LLMs reveal increasing application failures that persist even when the required knowledge is provided. 
Collectively, our findings highlight the value of knowledge-matched contrasts for diagnosing model limitations that conventional benchmark scores may obscure.

\newpage
\section*{Limitations} 
While \name{} enables more controlled evaluation of knowledge and application abilities, several limitations remain.

First, our framework currently focuses on multiple-choice questions, which facilitate automated answer validation and evaluation.
However, this restricts its applicability to open-ended tasks, where models must generate answers without predefined options.
Future work could extend \name{} to free-form question answering by incorporating semantic answer validation and evaluation methods that accommodate diverse valid responses.

Second, our proxy-model difficulty screening applies the same acceptance criterion to A2 and A3 candidates, limiting fine-grained control over question difficulty.
Moreover, because candidates are selected based on the performance of a specific proxy model, the retained questions may reflect model-specific weaknesses.
Although evaluations across diverse LLM families demonstrate diagnostic utility beyond the screening model, potential selection bias remains. 
Future work could introduce level-specific thresholds based on repeated proxy evaluations, such as requiring different minimum numbers of incorrect responses across multiple runs, to enable more flexible difficulty calibration.
Additionally, incorporating multiple proxy models or alternative screening configurations could reduce dependence on individual model capabilities while preserving the intended application requirements. 

Lastly, our experiments focus on questions derived from three existing Korean benchmarks.
While these benchmarks span diverse knowledge domains, the generalizability of \name{} to other languages and domains remains to be established.
Future research could extend the framework to a broader range of benchmarks to assess its applicability across different evaluation settings.

\section*{Broader Impact and Ethical Implications}
\name{} can contribute to more interpretable LLM evaluation on Korean benchmarks by distinguishing limitations in knowledge access from those in knowledge application.
Such diagnostic insights may support more targeted model development and reduce reliance on aggregate accuracy as the sole measure of model capability.

However, LLM-generated questions and automated judgments may inherit biases or inaccuracies from source benchmarks and underlying models, potentially affecting the fairness of evaluation.
Therefore, results from \name{} should complement broader evaluations rather than serve as definitive assessments of model capabilities, particularly in high-stakes settings.
In addition, since our framework constructs questions from existing benchmarks, the redistribution of generated materials should respect the licensing and attribution requirements of the original datasets.
We encourage transparent reporting of construction procedures and validation outcomes to support responsible use of the resulting benchmark.

\section*{Acknowledgments}

\noindent\textbf{Dataset Contributors.}
The KLEAR benchmark was validated by the following annotators at DATUMO INC., who assessed each generated question against the human rubric and provided the feedback used to revise it:

\noindent
Hosik Lee, Minseung Song, Jinsol Lee, Chanho Lee, Dahun Han, Nakyung Kim, Nawon Kim, Dongchan Lee, Dahye Kim, Hyeonah Ju, and Younghyun Jo.

%% file: 8_appendix.tex
\newpage

%%%%%%%%%%%%%%%%%%%% A. Source Benchmarks %%%%%%%%%%%%%%%%%%%%
\section{Source Benchmarks}
\label{app:source_benchmarks}

We use three existing Korean benchmarks, CLIcK, KMMLU-Pro, and KMMLU-Redux, to examine the knowledge and application requirements of existing evaluation questions.
This section describes the source benchmarks, data preparation, dataset statistics, and usage conditions.

\subsection{Benchmark Descriptions}

\paragraph{CLIcK.}
CLIcK~\citep{kim2024click} evaluates Korean cultural and linguistic knowledge using 1,995 multiple-choice questions collected from official examinations and educational materials.
It comprises 11 subcategories under two broad categories: Korean culture and Korean language.
These cover topics ranging from history, society, and law to grammar, textual knowledge, and language use.

\paragraph{KMMLU-Pro.}
KMMLU-Pro~\citep{hong-etal-2025-kmmlu} consists of 2,822 questions from Korean national professional licensure examinations.
It covers 14 professional qualifications across specialized domains, including law, taxation, accounting, and medicine.
The benchmark evaluates domain-specific knowledge required for professional qualifications.

\paragraph{KMMLU-Redux.}
KMMLU-Redux~\citep{hong-etal-2025-kmmlu} is a refined version of KMMLU, comprising 2,587 questions from Korean national technical qualification examinations.
It addresses issues identified in the original benchmark, including ambiguous questions, incorrect answers, and answer leakage.
The resulting benchmark covers technical and industrial knowledge across a range of specialized domains.

\subsection{Data Preparation}

We use all publicly released questions from the three benchmarks without subsampling, as distributed on the Hugging Face Hub (\texttt{EunsuKim/CLIcK}, \texttt{LGAI-EXAONE/KMMLU-Pro}, and \texttt{LGAI-EXAONE/KMMLU-Redux}; accessed on June 9, 2026).
We organize all questions into a common format containing the question, answer options, and gold answer, while retaining the source benchmark information for dataset-specific analyses.
For CLIcK, we prepend the accompanying passage to the question when available.
For KMMLU-Pro and KMMLU-Redux, we convert numeric answer indices into option letters.

Of the 7,404 source questions, 7,340 (99.1\%) are retained for our knowledge and application analysis.
The remaining 64 questions (62 from CLIcK and one each from KMMLU-Pro and KMMLU-Redux) are excluded due to errors encountered during the annotation process, rather than content-based filtering.

For contrast construction, we select 243 source questions as seeds for generating knowledge-matched contrasts, 165 of which yield the questions in the final benchmark.
The seed selection procedure and subsequent construction stages are described in Appendix~\ref{app:construction_details}.

\subsection{Dataset Statistics}
\input{Tables/appendix/source_stats}
Table~\ref{tab:source_stats} summarizes the original and retained question counts for each benchmark.
The resulting dataset spans diverse cultural, linguistic, professional, and technical knowledge domains.

\subsection{Data Sources, Licenses, and Usage Terms}
We use the publicly available versions of CLIcK, KMMLU-Pro, and KMMLU-Redux and cite their original publications.
KMMLU-Pro and KMMLU-Redux are distributed under the Creative Commons Attribution-NonCommercial-NoDerivatives 4.0 (CC BY-NC-ND 4.0) license, while the publicly available CLIcK documentation does not specify an explicit dataset license.

Since \name{} constructs questions derived from existing benchmark items, we contacted the creators of the source benchmarks to clarify their usage and redistribution conditions.
The authors of CLIcK permitted non-commercial use of both the original questions and modified versions.
LG AI Research permitted the modification and redistribution of KMMLU-Pro and KMMLU-Redux within the scope of its rights over the benchmarks as compilations and databases.
This permission does not extend to the individual examination items, whose rights remain with their original holders.

Accordingly, we release \name{} for non-commercial research and evaluation purposes.
The release scope and intended use of the constructed benchmark are further discussed in Appendix~\ref{app:benchmark_documentation}.

%%%%%%%%%%%%%%%%%%%% B. Knowledge and Application Annotation %%%%%%%%%%%%%%%%%%%%
\section{Knowledge and Application Annotation}
\label{app:ka_annotation}

We annotate the 7,340 retained questions from the three source benchmarks along two dimensions: the knowledge required to answer a question (K) and the application demand imposed by the question (A).
These dimensions are assessed separately to characterize whether a question primarily requires factual recall or the application of knowledge to given conditions.
This section describes the annotation taxonomy, silver reasoning generation, annotation procedure, and reliability assessment.

\subsection{Knowledge and Application Taxonomy}

We define four levels for each dimension, as summarized in Table~\ref{tab:ka_taxonomy}.

\input{Tables/appendix/ka_taxonomy}

\paragraph{Knowledge demand (K).}
Knowledge demand captures the level of prior knowledge required to answer a question.
The levels range from K0, where no external knowledge is required, to K3, where specialized professional, institutional, or procedural knowledge is necessary.
Intermediate levels distinguish everyday common sense and general language knowledge (K1) from school-level factual or conceptual knowledge (K2).
The assigned level reflects the knowledge required for a valid solution rather than the model's familiarity with the subject.

\paragraph{Application demand (A).}
Application demand captures how the required knowledge must be connected to the specific case in the question.
A0 questions can be answered through direct recall, since the retrieved fact or rule is itself the answer.
In A1 questions, the correspondence between the applicable knowledge and the case is stated explicitly.
In A2 questions, a single piece of knowledge applies, but one must judge whether the details of the case satisfy its condition.
A3 questions involve two or more competing pieces of knowledge or rules, and one must decide which of them governs the case.
Checking individual answer options one by one is not treated as evidence of A3, since it reflects the multiple-choice format rather than competition between candidate rules.
These levels describe the structure of the required application rather than empirical question difficulty.

Knowledge and application demand are treated as independent dimensions.
Tables~\ref{tab:k_examples} and~\ref{tab:a_examples} illustrate each level with a question from CLIcK, shown in Korean with our English translation.
A full example in which the same knowledge is held fixed across application levels is given in Appendix~\ref{app:qualitative}.

\input{Tables/appendix/k_examples}
\input{Tables/appendix/a_examples}

A question requiring specialized knowledge may involve only direct recall, while a question based on familiar knowledge may require more complex application.
This separation allows us to examine their distributions and associated model failures without treating knowledge specialization as a proxy for application demand.

\subsection{Silver Reasoning Generation}
\label{app:silver}

Determining the knowledge and application requirements of a question can be difficult from the question and gold answer alone, particularly when intermediate reasoning steps are implicit.
To provide additional reference information, we generate \textit{silver reasoning} for each question using GPT-5.4 with the temperature set to 0.

The generation model receives the question, answer options, and gold answer, and is instructed to produce a rationale explaining how the answer can be derived.
Providing the gold answer guides the generation toward the intended solution, rather than requiring the model to independently predict the correct option.

We refer to these rationales as \textit{silver} rather than gold reasoning because they are automatically generated and may not represent the only valid solution.
They serve as supplementary reference material for identifying the knowledge and application requirements of a question, rather than as independently verified ground-truth reasoning.
The full generation prompt is provided in Appendix~\ref{app:prompts}.

\paragraph{Answer recoverability.}
To check that silver reasoning contains sufficient information to derive the gold answer, we sample 150 questions (50 per benchmark), stratified by the number of the three Korean LLMs (Appendix~\ref{app:source_failure}) that answer the question correctly.
We remove explicit statements of the final answer from each rationale (218 statements in total) and ask GPT-5.4 to answer given only the question, the options, and the remaining reasoning.
The gold answer is recovered for all 150 questions (95\% Wilson interval: 97.5--100\%), compared with 118 (78.7\%) when the same model receives only the question and options.

\subsection{Annotation Procedure}

We use an LLM-based annotation procedure to assign K and A labels to the source questions, using GPT-5.4 with the temperature set to 0.
For each question, the annotator receives the question text, answer options, gold answer, and generated silver reasoning.
It then assigns one knowledge level (K0--K3) and one application level (A0--A3) according to the predefined taxonomy.
The annotator also assigns each question a content domain from a fixed list of ten domains, which we use to hold the domain fixed during construction (Appendix~\ref{app:construction_details}) and to describe the coverage of the benchmark (Appendix~\ref{app:benchmark_documentation}).

The annotator is instructed to judge the knowledge level first and then to judge the application level independently, based on how much work the silver reasoning needs to connect the knowledge to the case.
K and A are thus assessed as distinct dimensions, rather than inferred from the apparent difficulty of the question.
Silver reasoning is provided as supporting evidence to make implicit requirements more explicit.

The annotation procedure is applied consistently across CLIcK, KMMLU-Pro, and KMMLU-Redux.
Questions for which valid annotations could not be obtained are excluded, resulting in the 7,340 retained questions described in Appendix~\ref{app:source_benchmarks}.
The complete annotation instructions and prompt are provided in Appendix~\ref{app:prompts}.

\subsection{Annotation Reliability}

To assess the reliability of the K/A annotations, we conduct a human evaluation on 60 questions sampled evenly across the four application levels assigned by the LLM annotator (15 per level; 26 from CLIcK, 29 from KMMLU-Pro, and 5 from KMMLU-Redux).
Two human annotators independently assign K and A labels using the question, answer options, and gold answer, without access to the LLM-assigned labels or generated silver reasoning.
Both annotators are native Korean speakers with graduate-level training in NLP.
One is an author of this paper, while the other is not otherwise involved in the study.
Neither annotator received compensation for this evaluation.

We compare the human annotations with LLM-generated labels under two conditions: with and without silver reasoning.
The condition without silver reasoning uses the same annotation prompt with the silver reasoning removed.
Since both dimensions are ordinal, we report linearly weighted Cohen's $\kappa$, along with exact agreement.
LLM--human agreement is computed separately against each annotator and averaged over the two.
Table~\ref{tab:ka_reliability} presents the results.

\input{Tables/appendix/ka_reliability}

For application demand, LLM--human agreement reaches $\kappa=0.69$ with silver reasoning, compared with $\kappa=0.47$ without it.
The former also exceeds the human--human agreement of $\kappa=0.60$, suggesting that silver reasoning provides useful supplementary information for identifying application requirements that may be implicit in the question alone.

For knowledge demand, however, silver reasoning does not improve agreement.
LLM--human agreement is $\kappa=0.56$ with silver reasoning and $\kappa=0.67$ without it, compared with human--human agreement of $\kappa=0.63$.
In particular, the frequency of K2 assignments differs substantially: the LLM with silver reasoning assigns K2 to 21 questions, whereas the two human annotators assign it to 14 and 5 questions, respectively.
These differences suggest that knowledge-level annotation is more sensitive to annotator calibration, particularly at intermediate levels, and that the benefits of silver reasoning do not extend uniformly across both dimensions.

%%%%%%%%%%%%%%%%%%%% C. Failure Analysis on Source Benchmarks %%%%%%%%%%%%%%%%%%%%
\section{Failure Analysis on Source Benchmarks}
\label{app:source_failure}

We analyze incorrect responses to the three source benchmarks to investigate how model failures vary with application demand.
This section describes the failure taxonomy, classification procedure, and observed failure patterns.

\subsection{Failure Taxonomy}

We categorize incorrect responses into knowledge failures, application failures, and other errors.
Knowledge failures arise when the model lacks or incorrectly recalls information required to answer the question.
Application failures occur when the required knowledge is available but the model fails to apply it correctly to the given conditions.
Other errors include invalid outputs and defective questions for which a reliable attribution cannot be established.

The fine-grained taxonomy consists of 13 failure types (T0--T12), listed in Table~\ref{tab:failure_taxonomy}.
T0--T2 cover invalid outputs and defective questions, T3--T6 cover knowledge failures, and T7--T12 cover application failures.
These types support a more detailed characterization of model errors while allowing aggregation into the three broader groups.

\input{Tables/appendix/failure_taxonomy}

\subsection{Failure Classification Procedure}

We first collect incorrect closed-book responses from three Korean LLMs: K-EXAONE, A.X K1, and Solar Open.
Across the 7,340 retained questions, this yields 5,808 incorrect responses (1,859 from K-EXAONE, 1,681 from A.X K1, and 2,268 from Solar Open).

We use an LLM-based classifier, GPT-5.4 with the temperature set to 0, to assign failure types according to the predefined taxonomy.
For each question, the classifier receives the question, answer options, gold answer, and silver reasoning, together with the extracted answer and reasoning trace of each model that answered incorrectly.
It assigns one or more failure types to each incorrect response, ordered from the most to the least influential.
Most responses receive a single type (4,941 of 5,808), and we use the first type as the primary failure type in all analyses.

We additionally examine model behavior when the required knowledge is supplied, as supporting evidence for the failure attribution.
On a subset of 702 questions with knowledge or application failures, we provide K-EXAONE and Solar Open with gold knowledge and re-evaluate them.
Responses whose primary failure is incorrect recall (T3) are corrected in 91.0\% and 84.7\% of cases, respectively, compared with 75.9\% and 78.9\% for responses that misjudge the scope or condition of a correctly stated rule (T10).
Errors attributed to knowledge are thus recovered more often than errors attributed to application, although most of the latter are also corrected.
We use these observations as diagnostic evidence rather than definitive explanations of individual model errors.

\subsection{Failure Patterns Across Application Levels}

Table~\ref{tab:failure_by_a} reports the composition of failure types at each application level.
Recall-only questions (A0) account for the largest portion of existing benchmark questions (4,482 of 7,340; 61.1\%), and incorrect responses at this level predominantly reflect knowledge failures (69.8\%).

\input{Tables/appendix/failure_by_a}

As application demand increases, application failures account for a larger share of incorrect responses, rising from 21.1\% at A0 to 36.6\% at A1, 41.9\% at A2, and 44.0\% at A3.
At the same time, knowledge failures remain about as frequent as application failures at A2 and A3.
This suggests that questions requiring more complex application expose limitations that are less apparent in recall-oriented evaluation, while still depending on whether the model holds the required knowledge.

Comparisons across existing questions therefore cannot isolate the effect of application demand, since their underlying knowledge requirements also differ.
This motivates the knowledge-matched contrasts constructed by \name{}, which enable more controlled comparisons while preserving the required knowledge.

%%%%%%%%%%%%%%%%%%%% D. KLEAR Construction Details %%%%%%%%%%%%%%%%%%%%
\section{\name{} Construction Details}
\label{app:construction_details}

This section describes each stage of the construction pipeline: seed selection and anchor knowledge extraction, contrast generation, LLM-based auditing, proxy-model difficulty screening, and revision.
Stage-wise yields and human validation are reported in Appendix~\ref{app:construction_quality}.

\subsection{Anchor Knowledge Extraction}

\paragraph{Seed selection.}
Seeds are source questions labeled A1 or A2 in Appendix~\ref{app:ka_annotation}.
An A1 seed can be transformed into A2 and A3 contrasts, and an A2 seed into an A3 contrast, so that a contrast always has a higher application level than its seed.
We select questions that at least two of the three Korean LLMs answer incorrectly, which holds for 237 of the 243 seeds; the remaining six are A1 questions chosen in a pilot study for their suitability for transformation.
The 243 seeds comprise 78 A1 and 165 A2 questions (88 from CLIcK, 84 from KMMLU-Pro, and 71 from KMMLU-Redux).

\paragraph{Anchor bundle.}
For each seed, GPT-5.5 extracts an \textit{anchor bundle} from the question, answer options, gold answer, and silver reasoning.
The bundle is a closed set of the knowledge on which the question rests and consists of three parts: (i) \textit{knowledge units}, each a self-contained statement of a rule, criterion, definition, or fact; (ii) \textit{relations} that describe how the units interact, such as which condition decides between two units or which unit is the principle and which the exception; and (iii) a \textit{scope note} stating what the bundle excludes.
The knowledge units cover the knowledge behind the distractors as well as the gold answer, since competing knowledge is what makes an A3 contrast possible.
A bundle contains 4 to 17 knowledge units (median 8) and 4 to 9 relations (median 6).

\paragraph{Transformability.}
The same model also judges whether the bundle can support each target level.
A seed supports A2 if its stated correspondence can be hidden so that the condition of the applicable knowledge must be inferred from the case, and supports A3 if two or more knowledge units could each plausibly govern the same case.
We generate contrasts only for supported targets, which yields 295 seed--target pairs: 78 from A1 to A2, 52 from A1 to A3, and 165 from A2 to A3.

\subsection{Application-Controlled Contrast Generation}

For each seed--target pair, GPT-5.5 generates candidate questions given the seed question with its options, gold answer, and silver reasoning, the knowledge level and domain of the seed, the anchor bundle, and the target-application level.
The prompt includes few-shot demonstrations of the target level.

The generator is instructed to keep the anchor bundle, knowledge level, and domain fixed and to change only the application demand.
An A2 candidate must involve exactly one applicable piece of knowledge whose condition is not stated but must be inferred from the case.
An A3 candidate must present a single case for which two or more pieces of knowledge in the bundle compete, with case facts of comparable weight supporting each.
In both cases, the candidate must ask for the same kind of answer as the seed, must not state or paraphrase the anchor knowledge, and must not require knowledge outside the bundle.

Each candidate has four or five options.
Every distractor must correspond to one identifiable error, either applying a competing piece of knowledge or misjudging the condition of the correct one, and no option may be eliminable on grounds that bypass the intended application.
Along with the question, the generator outputs its reasoning, an account of the knowledge-to-case mappings involved, and an account of the error behind each distractor.

We first generate three candidates per pair.
A pair that does not obtain two passing candidates after auditing, screening, and revision receives six additional candidates.
This results in 1,964 candidates in total.

\subsection{LLM-Based Auditing}
\input{Tables/appendix/rubrics}
Each candidate is audited by GPT-5.5 against the eleven criteria of the LLM rubric (L1--L11; Table~\ref{tab:rubrics}).
The auditor receives the seed question, the anchor bundle, the target-application level, and the candidate with its proposed gold answer.
It does not receive the reasoning or accounts written by the generator, and is instructed to solve the candidate independently before judging it.

For each criterion, the auditor returns \textsc{pass}, \textsc{fail}, or \textsc{uncertain} together with a rationale.
A candidate satisfies the rubric only if all eleven criteria are \textsc{pass}; \textsc{uncertain} is treated as a failure.
For shortcut resistance (L8), the auditor additionally reports, for each distractor, the least demanding basis on which it can be eliminated.

\subsection{Proxy-Model Difficulty Screening}

A candidate can satisfy the rubric and still be answerable without the intended application.
We therefore screen candidates with a proxy model, GPT-5-nano, which is considerably weaker than the models used for generation and auditing.
The proxy receives only the question and options, without the anchor bundle, the seed, or any audit result, so that the screening is independent of the audit.
A candidate passes the automated stage only if it satisfies all eleven criteria and the proxy answers it incorrectly.

\subsection{Iterative Revision}

\paragraph{Self-revision.}
Candidates that do not pass the automated stage are revised once by GPT-5.5.
The reviser receives the candidate, the rationale of every criterion that is not \textsc{pass}, and, when applicable, the fact that the proxy answered the candidate correctly.
It is instructed to resolve the listed defects without regressing other criteria, to introduce no knowledge outside the anchor bundle, and to keep the knowledge level, domain, and target-application level fixed.
When a candidate has no listed defect but is answered correctly by the proxy, the reviser is instructed to locate and close the shortcut rather than alter the question otherwise.
A revised candidate is treated as a new question: it is audited and screened again under the same conditions.

\paragraph{Human-guided revision.}
Candidates that pass the automated stage are forwarded to human validation, with at most two candidates per pair.
Candidates that fail human validation are revised by GPT-5.5 using the rationales written by the annotators and are then validated again (Appendix~\ref{app:construction_quality}).

\paragraph{Final questions.}
The final 287 questions cover 195 seed--target pairs and 165 seeds, with one question for 103 pairs and two for 92.
Of these, 182 have four options and 105 have five.
We permute the option order of 192 questions to balance the position of the gold answer (A: 68, B: 68, C: 66, D: 65, E: 20).

\subsection{Construction Configurations}

Table~\ref{tab:construction_config} lists the model used at each stage.
All stages use the default sampling settings of the API and structured outputs.

\input{Tables/appendix/construction_config}

%%%%%%%%%%%%%%%%%%%% E. Construction Quality and Validation %%%%%%%%%%%%%%%%%%%%
\section{Construction Quality and Validation}
\label{app:construction_quality}

This section reports the yield of each construction stage, the human validation protocol and its results, and additional analyses of the anchor bundle and seed difficulty.

\subsection{Stage-Wise Construction Statistics}

Table~\ref{tab:yield_by_type} breaks down the yield of the automated stage by transformation type.
Before revision, the share of candidates that satisfy all eleven criteria is similar across types (30.7--36.5\%), whereas the share that the proxy answers incorrectly differs: 25.8\% for A1$\rightarrow$A2, 36.2\% for A1$\rightarrow$A3, and 45.8\% for A2$\rightarrow$A3.
A2 contrasts are thus screened out more often for being answerable by the proxy than for failing the rubric.
Self-revision recovers 17.2--21.6\% of the revised candidates in each type.

\input{Tables/appendix/yield_by_type}

Yield also differs across source benchmarks.
The share of candidates that pass the automated stage is 24.2\% for seeds from CLIcK (188 of 777), 43.0\% for KMMLU-Pro (259 of 602), and 30.1\% for KMMLU-Redux (176 of 585).

\subsection{Human Validation Protocol}

\paragraph{Candidates.}
Human validation covers 301 of the 623 candidates that pass the automated stage.
In the first batch, up to two passing candidates per pair are validated, which gives 231 candidates.
In the second batch, because of time constraints, one candidate is validated for 70 of the 111 pairs that have a passing candidate.

\paragraph{Procedure.}
Annotators judge each candidate against the five criteria of the human rubric (H1--H5; Table~\ref{tab:rubrics}) in two stages.
In the first stage, they see only the question, options, and gold answer, and judge Korean readability (H1), natural construction (H2), and surface leakage (H3).
They are instructed not to verify the answer with domain knowledge at this stage.
After these judgments are recorded, the anchor bundle is revealed, and annotators judge knowledge sufficiency (H4) and target-level structure (H5).

Each criterion is rated on a three-level scale, and a criterion is satisfied only at the highest level.
Any lower rating must be accompanied by a written rationale, which is later used for revision.
A candidate passes human validation when it satisfies all five criteria.
When a candidate is judged by two annotators, it fails if either annotator rates any criterion below the highest level.

\paragraph{Annotators.}
All annotators are native Korean speakers who work as professional data annotators and follow written guidelines calibrated on a pilot set, as described in Sec.~\ref{sec:construction_setup}.
None of them is an author of this paper, and they received no additional compensation for this evaluation.

\paragraph{Revision.}
Candidates that do not pass are revised by GPT-5.5 as described in Appendix~\ref{app:construction_details}, using the rationales written by the annotators, and the revised candidates are validated again under the same rubric.

\subsection{Detailed Validation Results}

Table~\ref{tab:human_validation} summarizes the outcome of human validation for each batch.
Of the 301 candidates, 252 (83.7\%) pass directly.
The remaining 49 are revised, and 35 of them (71.4\%) pass after revision, while 14 are discarded.
This yields the final 287 questions.

\input{Tables/appendix/human_validation}

Table~\ref{tab:criteria} reports failures by criterion for both rubrics.
Under the human rubric, failures are spread across criteria rather than concentrated in one: 7.0\% of the candidates fail Korean readability (H1), 5.0\% surface leakage (H3), 4.3\% target-level structure (H5), 3.3\% natural construction (H2), and 2.0\% knowledge sufficiency (H4).

\input{Tables/appendix/criteria}

\subsection{Additional Construction Ablations}

\paragraph{Effect of the anchor bundle.}
To examine the role of the anchor bundle, we regenerate candidates for 36 seed--target pairs without it, producing three candidates per pair (108 in total), and compare them with the 108 candidates generated with the bundle for the same pairs.
In the condition without the bundle, the generator is still instructed to keep the knowledge of the seed fixed, but must identify that knowledge from the seed question and its silver reasoning.
All other parts of the prompt, including the level definitions, option rules, and demonstrations, are unchanged, so the two conditions differ only in whether the bundle is given.
Both sets are audited and screened under the same conditions as in the main pipeline, and we compare them before revision (Table~\ref{tab:anchor}).

\input{Tables/appendix/anchor}

Without the bundle, 20.4\% of the candidates require knowledge outside the anchor bundle (L4), compared with 1.9\% with the bundle (Fisher's exact test, $p < 0.001$).
No other criterion differs significantly, and the share of candidates answered correctly by the proxy is nearly identical (51.9\% vs.\ 50.0\%).

\paragraph{Effect of seed difficulty.}
\label{app:easy}
Most seeds in the main pipeline are questions that at least two of the three Korean LLMs answer incorrectly.
To examine how this choice affects yield, we additionally process 20 A1 seeds that are easier for these models, 14 answered incorrectly by one model and 6 by none, with the same pipeline.
Candidates from these seeds are used only for this analysis and are not included in \name{}.
Table~\ref{tab:seed_difficulty} groups all A1 seeds by the number of models that answer the seed incorrectly.

\input{Tables/appendix/seed_difficulty}

Candidates from easier seeds satisfy the LLM rubric more often before revision: 58.1\% for seeds that no model answers incorrectly, compared with 28.0\% for seeds that all three answer incorrectly.
However, they are much less likely to be answered incorrectly by the proxy (12.8\% vs.\ 39.8\%).
As a result, the share of candidates that pass the automated stage is lower for easier seeds, 18.8\% and 21.5\% for seeds with zero and one incorrect model, compared with 29.9\% and 28.9\% for seeds with two and three.
Seeds that are easy for current models thus tend to yield well-formed contrasts that remain easy, which supports selecting seeds on which models fail.

%%%%%%%%%%%%%%%%%%%% F. Constructed Benchmark Documentation %%%%%%%%%%%%%%%%%%%%
\section{Constructed Benchmark Documentation}
\label{app:benchmark_documentation}

This section documents the final benchmark: its statistics, coverage and intended use, privacy and content considerations, and release conditions.

\subsection{Final Benchmark Statistics}

\name{} consists of 287 multiple-choice questions in Korean, 78 at A2 and 209 at A3, derived from 165 seeds.
Table~\ref{tab:final_stats} reports their composition by application level.
Most questions require school-level (K2; 41.5\%) or specialized (K3; 47.0\%) knowledge, following the knowledge levels of their seeds.
A3 questions are longer than A2 questions, with a mean of 349 characters against 198, since they must present case facts that support two or more competing pieces of knowledge.

\input{Tables/appendix/final_stats}

Each seed has at most two questions per target level.
Thirty seeds have both an A2 and an A3 contrast, which allows the two levels to be compared on the same underlying knowledge.

\subsection{Dataset Coverage and Intended Use}

\paragraph{Coverage.}
We describe coverage with the content domain assigned to each source question during annotation (Appendix~\ref{app:ka_annotation}).
The questions cover language and culture (105), medicine and health (31), economy and finance (30), law and administration (27), history and society (19), science and technology (18), education (4), labor and employment (2), and other specialized domains (51).
The 130 questions derived from CLIcK come mostly from its language category (105), covering grammar (50), textual (28), and functional (27) knowledge, with 25 from the culture category.
Coverage follows from which seeds support a transformation and survive validation, and is therefore not balanced across domains or source benchmarks.

\paragraph{Intended use.}
\name{} is intended for diagnostic evaluation of Korean LLMs: comparing performance across application levels while the required knowledge is held fixed, and separating failures of knowledge from failures of application.
It is designed for evaluation only and should not be used as training data.
It is not intended to certify professional competence in the domains it covers, although many of its seeds originate from licensure examinations.

\paragraph{Limitations.}
The benchmark is small and unbalanced across levels, and only 30 seeds support a direct comparison between A2 and A3.
Seeds are selected among questions that Korean LLMs answer incorrectly, so the questions are not a representative sample of the source benchmarks.
All questions are generated and revised by a single model family and audited automatically before human validation, and the anchor bundles are generated automatically as well.
Finally, the questions follow a multiple-choice format, which constrains how application demand can be expressed.

\subsection{Privacy and Content Review}

The seeds come from public examinations and educational materials, and the generated questions present cases written for the question.
We screened all 287 questions and their options for e-mail addresses, telephone numbers, resident registration numbers, card numbers, and URLs, and found none.
Every question in the final benchmark was also read in full by at least one human annotator during validation.
No privacy or offensive-content concerns were raised during validation.

\subsection{Data Release and Licensing}

We release the 287 generated questions for non-commercial research and evaluation purposes, under the permissions described in Appendix~\ref{app:source_benchmarks}.
Each question is released with its options, gold answer, target-application level, and the source benchmark and item ID of its seed.
The source questions are not redistributed and must be obtained from their original distributions under their own terms.
Of the 287 questions, 130 are derived from 66 CLIcK seeds, 90 from 53 KMMLU-Pro seeds, and 67 from 46 KMMLU-Redux seeds.
\name{} is distributed solely by the authors of this work and is not a joint release with the creators of the source benchmarks.
Following the source benchmarks, the data should not be included in training corpora.

%%%%%%%%%%%%%%%%%%%% G. Experimental Setup %%%%%%%%%%%%%%%%%%%%
\section{Experimental Setup}
\label{app:experimental_setup}

This section describes the evaluated models, inference configurations, evaluation conditions and metrics, and the computational resources used.

\subsection{Evaluated Models}

We evaluate eight models, listed in Table~\ref{tab:eval_models}: three Korean LLMs (K-EXAONE, A.X K1, and Solar Open), a multilingual model (Aya Expanse 32B), two open-weight models (GLM-4.7 and Qwen3.5), and two proprietary models (Claude Sonnet 5.5 and GPT-6.1 Sol).
The three Korean LLMs are evaluated on both the 287 questions of \name{} and their 165 seeds.
The other five models are evaluated on the 287 questions of \name{}.

\input{Tables/appendix/eval_models}

\subsection{Inference Configurations}

\paragraph{Serving and decoding.}
The three Korean LLMs are served locally with vLLM, with reasoning enabled and the sampling settings in Table~\ref{tab:eval_models}.
The remaining models are accessed through APIs.
GLM-4.7 and Qwen3.5 are accessed through OpenRouter, with each model pinned to the endpoint of its developer so that all calls are served by the same deployment.
Reasoning is set to high effort for every model that supports it; Aya Expanse 32B has no reasoning mode.
API models use their default sampling settings, except Aya Expanse 32B, for which the temperature is set to 0.3.

\paragraph{Prompt.}
Each question is sent once as a single user message without a system message.
The message consists of an instruction, the question, and the options, each labeled with a letter.
We use the instruction of the source benchmark from which the seed originates, which asks the model to reason step by step and to give its final answer after the marker ``정답:'' (``Answer:'').
The prompt and all other settings are identical across evaluation conditions, apart from the added context described below.

\paragraph{Answer extraction.}
We take the letter in the last occurrence of the answer marker in the final response.
If the final response contains no marker, we apply the same rule to the reasoning output, and otherwise treat the response as having no answer.
Responses with no answer, including those truncated at the output limit, are scored as incorrect.

\subsection{Evaluation Conditions and Metrics}

\paragraph{Conditions.}
We evaluate each question under three conditions.
The \textit{Closed} (closed-book) condition provides only the question and options.
The \textit{+K} condition additionally provides the knowledge required by the question, as three knowledge statements.
The \textit{+K+A} condition provides the same three statements together with statements on how the knowledge applies (between 2 and 9, with a median of 5).
The added context is inserted between the question and the options under a neutral header (``참고 자료'', ``Reference material''), and the model is not told which condition it is in.

The context is generated by GPT-5.4 from the anchor bundle of each seed, not from individual questions.
A seed and all contrasts derived from it therefore receive exactly the same context, so that only the application level differs between them.
The median context length is 276 characters for \textit{+K} and 769 characters for \textit{+K+A}.

\paragraph{Metrics.}
We report accuracy, the share of questions answered correctly.
For the comparison between A2 and A3 on the same knowledge, we use the 30 seeds that have both an A2 and an A3 contrast and average accuracy over seeds.
Failure types on \name{} are assigned with the classifier of Appendix~\ref{app:source_failure}.

\paragraph{Unanswered responses.}
On the 861 calls for \name{} (287 questions under three conditions), K-EXAONE, A.X K1, and GLM-4.7 reach the output limit in 26, 30, and 7 calls, respectively, and no other model does. Responses without an extracted answer number 25 for K-EXAONE, 25 for A.X K1, 7 for Solar Open, 6 for GLM-4.7, two for Claude Sonnet 5.5, one for Qwen3.5, and none for GPT-6.1 Sol.
Aya Expanse 32B yields 134 such responses (15.6\%), which should be taken into account when interpreting its accuracy.

\subsection{Computational Resources and Costs}

The three Korean LLMs are served with tensor parallelism over four (K-EXAONE), eight (A.X K1), and two (Solar Open) GPUs.

Table~\ref{tab:usage} reports the token usage of the API calls for construction and evaluation.
Construction with GPT-5.5 uses 34.4M input tokens and 65.0M output tokens across generation, auditing, and revision, and the proxy model uses a further 22.5M output tokens.
Evaluation is much cheaper: each API model processes the 861 calls with between 0.3M and 3.7M output tokens.
The costs reported by OpenRouter are \$7.75 for GLM-4.7 and \$8.85 for Qwen3.5.

\input{Tables/appendix/usage}

%%%%%%%%%%%%%%%%%%%% H. Extended Diagnostic Results %%%%%%%%%%%%%%%%%%%%
\section{Extended Diagnostic Results}
\label{app:extended_results}

This section extends the results on \name{} by source benchmark and model family, examines the effect of providing knowledge in more detail, analyzes failure types on the constructed contrasts, and reports statistical tests and robustness checks.

\subsection{Results by Source Benchmark}

Table~\ref{tab:by_benchmark} reports accuracy under the three conditions, grouped by the source benchmark of the seed.
In the \textit{Closed} condition, the three Korean LLMs are least accurate on questions derived from KMMLU-Pro (23.3--40.0\%), which require specialized professional knowledge.
Providing knowledge raises accuracy on all three groups, and the gain is largest for questions derived from KMMLU-Pro and KMMLU-Redux.
With both knowledge and application statements, the Korean LLMs reach 88.1--95.5\% on questions derived from KMMLU-Redux but only 74.6--86.9\% on those derived from CLIcK.
The questions derived from CLIcK, which mostly concern language, thus remain the hardest once the required knowledge and its application are supplied.

\input{Tables/appendix/by_benchmark}

\subsection{Results by Model Family}

Table~\ref{tab:by_family} averages accuracy within four groups of models.
The ordering of the groups is the same at both levels and under all conditions: proprietary models are the most accurate, followed by open-weight models, Korean LLMs, and the multilingual model.
In every group, accuracy at A3 is lower than at A2, and the difference widens when knowledge is provided, from 12.1 to 19.1 points for Korean LLMs and from 4.6 to 7.5 points for proprietary models.
Supplying knowledge removes much of the difficulty at A2, whereas a larger part remains at A3.
We note that the questions are generated and audited with models of the same developer as GPT-6.1 Sol, which may favor that model.

\input{Tables/appendix/by_family}

\subsection{Effects of Knowledge Provision}

\paragraph{From knowledge to application statements.}
Since \textit{+K} and \textit{+K+A} are applied to the same questions, we compare them with paired tests (Table~\ref{tab:knowledge_gain}).
At A3, adding application statements corrects more responses than it breaks for every model, and the net gain is significant for seven of the eight models.
At A2, few errors remain once knowledge is provided, between 0 and 13 for all models other than Aya Expanse 32B, and the gain is not significant for any model.
The errors that persist after knowledge is supplied are therefore concentrated at A3, and a substantial part of them is resolved by stating how the knowledge applies.

\input{Tables/appendix/knowledge_gain}

\paragraph{Results on seeds.}
Table~\ref{tab:seed_results} reports the accuracy of the three Korean LLMs on the 165 seeds under the same conditions.
The pattern mirrors that on the contrasts: with knowledge provided, accuracy is 81.0--84.5\% on A1 seeds and 60.7--73.8\% on A2 seeds.
Adding application statements yields a significant gain on A2 seeds for all three models ($p = 0.021$, $0.049$, and $0.004$), and on A1 seeds only for Solar Open ($p = 0.039$).
Accuracies on seeds and on contrasts are not directly comparable, because seeds are selected among questions that these models answer incorrectly and are not subject to the validation applied to the contrasts.

\input{Tables/appendix/seed_results}

\subsection{Failure Analysis on Constructed Contrasts}

We classify the incorrect \textit{Closed} responses of the three Korean LLMs on \name{} with the classifier of Appendix~\ref{app:source_failure}.
The setup is identical except for the reference solution: in place of silver reasoning, the classifier receives the reasoning written by the generator for the question.
It is not given the anchor bundle or the target-application level.
In total, 237 of the 287 questions have at least one incorrect response, giving 514 responses.

Table~\ref{tab:failure_contrast} reports the composition of primary failure types.
Application failures account for 67.2\% of the incorrect responses at A2 and 73.4\% at A3, and the pattern holds for each model (65.2--70.6\% at A2 and 71.8--74.6\% at A3).
For comparison, application failures account for 41.9\% and 44.0\% of the incorrect responses on source questions at A2 and A3 (Table~\ref{tab:failure_by_a}).
The dominant type is misjudging the scope or condition of a correctly stated rule (T10), with 70 of 119 responses at A2 and 218 of 395 at A3, followed by misapplying a structural rule (T9), with 9 and 50 responses.

\input{Tables/appendix/failure_contrast}

\subsection{Statistical Analysis and Robustness}

\paragraph{Tests.}
For paired comparisons between conditions, we use the exact McNemar test.
For the comparison between A2 and A3 on matched seeds, we compute accuracy per seed at each level and apply a two-sided sign test over the 30 seeds.
We do not correct for multiple comparisons, and the $p$-values should be read accordingly.

\paragraph{Matched comparison of A2 and A3.}
Table~\ref{tab:matched_test} reports the matched comparison in the \textit{Closed} condition.
For all eight models, more seeds favor A2 than A3, but the difference is significant only for A.X K1 ($p = 0.019$).
With 30 seeds, of which many are tied for a given model, the test has limited power, and we treat the consistent direction across models as suggestive rather than conclusive.
We rely on matched seeds for this comparison because, across all questions, the two levels are built from different seeds and their difference is confounded with seed difficulty.

\input{Tables/appendix/matched_test}

\paragraph{Unanswered responses.}
Responses without an extracted answer are scored as incorrect, which could distort comparisons if they were unevenly distributed.
Across the 861 calls, such responses number 25 for K-EXAONE, 25 for A.X K1, 7 for Solar Open, 6 for GLM-4.7, 2 for Claude Sonnet 5.5, 1 for Qwen3.5, and none for GPT-6.1 Sol, compared with 134 for Aya Expanse 32B.
When these responses are excluded, the accuracy of Aya Expanse 32B rises by 4 to 11 points depending on the level and condition, while that of the other models changes by at most 3.4 points.
The difference between A2 and A3 in the \textit{Closed} condition changes by at most 1.7 points for any model.
The gain from application statements at A3 remains significant for six models when restricted to questions answered under both conditions, with Claude Sonnet 5.5 at $p = 0.064$ and GPT-6.1 Sol unchanged at $p = 0.109$.

%%%%%%%%%%%%%%%%%%%% I. Qualitative Analysis %%%%%%%%%%%%%%%%%%%%
\section{Qualitative Analysis}
\label{app:qualitative}

This section presents examples of a knowledge-matched contrast, of candidates that were rejected and revised, and of a model failure.
Korean text is shown with our English translation.

\subsection{Examples of Knowledge-Matched Contrasts}

Table~\ref{tab:example_contrast} shows an A2 and an A3 contrast built from the same seed, a question from KMMLU-Redux on methods for recovering accelerants from fire debris.
The seed is an A1 question: it names the procedure, suspending charcoal in the headspace of a container that holds residue, and asks which method this is.
The correspondence between the knowledge and the case is thus stated outright.

The A2 contrast describes the same procedure without naming the adsorbent, referring only to ``small black carbonaceous granules'' in a metal mesh.
A single piece of knowledge still applies, but whether the case meets its condition, that the components are captured on the surface of an adsorbent, must now be inferred.
The A3 contrast keeps the answer and the options but adds case facts that support the other methods in the bundle: a hot water bath and droplets inside the lid point to steam distillation, placing the carbon in a vial of solvent points to solvent extraction, and the space above the debris points to the headspace method.
Deciding which method governs the case is now the task.
All three questions rest on the same anchor bundle, and only the application demand differs.

\input{Tables/appendix/example_contrast}

\subsection{Examples of Rejected and Revised Questions}

\paragraph{Self-revision after automated auditing.}
Table~\ref{tab:example_selfrev} shows an A2 candidate that fails shortcut resistance (L8).
The question asks for which underground facility the largest excavation support ratio should be chosen, and the intended inference is that a facility in temporary use permits a larger ratio.
In the first version, the gold option is the only long and detailed one, while the others are bare names of facilities, so the auditor judges that the gold answer can be isolated from its form alone.
The revision rewrites all four options in the same form, each describing how the facility is planned to be used, so that the options must be compared on the intended criterion.

\input{Tables/appendix/example_selfrev}

\paragraph{Human-guided revision.}
Table~\ref{tab:example_humanrev} shows an A2 candidate that satisfies all eleven LLM criteria but fails surface leakage (H3) in human validation.
The question concerns which authority manages a road sign in a county that now belongs to a metropolitan city.
The stem states that the notation with the former province is an outdated one that remains on some notices, and the annotator points out that the two options naming that province can therefore be eliminated by reading alone, without any knowledge of the law.
The revision removes the sentence about the former notation and replaces the two options with authorities of the metropolitan city, so that no option is excluded by the stem itself.

\input{Tables/appendix/example_humanrev}
\input{Tables/appendix/example_failure}
\subsection{Representative Model Failures}

Table~\ref{tab:example_failure} shows the responses of A.X K1 to the A3 contrast of Table~\ref{tab:example_contrast}.
In the \textit{Closed} condition, the model recalls what each method is and recognizes both the adsorption onto activated carbon and the later transfer to a solvent, but concludes that the procedure as a whole is the headspace method because the vapor is collected in the space above the sample.
The classifier assigns this response to T10: the relevant rules are stated correctly, but their scope is misjudged.

Providing knowledge does not resolve the failure.
Under \textit{+K}, the model is given the definitions of all four methods and changes its answer to solvent extraction, another incorrect method that the case also supports.
Only under \textit{+K+A}, where the context states that the method is determined by the recovery principle rather than the location, and that capture on an adsorbent before transfer to a solvent counts as adsorption, does the model answer correctly.
The same pattern appears in aggregate: at A3, errors that persist after knowledge is provided are largely resolved once its application is stated (Appendix~\ref{app:extended_results}).

%%%%%%%%%%%%%%%%%%%% J. Prompts and Annotation Instructions (short version) %%%%%%%%%%%%%%%%%%%%

\section{Prompts and Annotation Instructions}
\label{app:prompts}

This section lists the prompts that define the annotation taxonomy, the construction of contrasts, and the evaluation, together with the instructions given to human annotators.
The remaining prompts, for failure classification, revision, and the generation of the context used in the \textit{+K} and \textit{+K+A} conditions, are provided in the supplementary material.
Prompts are reproduced as used, except that a few typographic symbols (dashes and arrows) are replaced by ASCII equivalents.
Unless noted otherwise, each prompt is given as the system message, and the user message is a JSON object with the inputs described in the corresponding appendix.

\subsection{Annotation Prompts}

\subsubsection{Prompt for Silver Reasoning Generation}

The user message contains the question, options, and gold answer (Appendix~\ref{app:ka_annotation}).

\promptfile{prompts/01_silver_reasoning.txt}

\subsubsection{Prompt for Knowledge and Application Annotation}

The user message contains the question, options, gold answer, and silver reasoning.
We show the part of the prompt that defines the K and A levels.
The remainder defines the content domain and auxiliary labels that are not used in this paper.

\promptfile{prompts/02a_ka_annotation_core.txt}

\subsection{Construction Prompts}

\subsubsection{Prompt for Anchor Knowledge Extraction}

We show the prompt for A1 seeds.
The prompt for A2 seeds differs only in its description of the seed and omits the judgment on whether an A2 contrast is supported.

\promptfile{prompts/04_anchor_extraction.txt}

\subsubsection{Prompt for Application-Controlled Contrast Generation}

The system message consists of the following contract, followed by the fixed attributes of the seed (knowledge level, domain, and target level) and by demonstrations of the target level, which we omit here.
We show the contract for A1 seeds.

\promptfile{prompts/05_generation_contract.txt}

\subsubsection{Criteria for LLM-Based Auditing}

The eleven criteria of the LLM rubric are passed to the auditor with the following wording, together with the seed question, the anchor bundle, the target level, and the candidate.

\promptfile{prompts/06b_auditing_rubric.txt}

\subsection{Evaluation Prompts}

Each question is sent as a single user message without a system message.
The instruction follows the source benchmark of the seed, and \texttt{\{letters\}} lists the option letters of the question.

\promptfile{prompts/08_model_evaluation.txt}

In the \textit{+K} and \textit{+K+A} conditions, the context is inserted between the question and the options under the header ``참고 자료'' (``Reference material'').

\promptfile{prompts/09_knowledge_provision.txt}

\subsection{Human Annotation Instructions}

Table~\ref{tab:human_rubric} gives the question and the three rating levels shown to annotators for each criterion of the human rubric.
In the first stage, the annotation interface shows only the question, options, and gold answer, with the instruction ``전문지식으로 정답을 검증하지 말고 보이는 문항만 평가합니다'' (``Do not verify the answer with expert knowledge; evaluate only the question as shown'').
After the ratings for H1--H3 are fixed, the interface reveals the anchor bundle with the instruction ``왼쪽의 anchor를 읽고 두 기준을 평가합니다'' (``Read the anchor on the left and evaluate the two criteria'').
A written note is required for every rating below the highest level.

\input{Tables/appendix/human_rubric}

%%%%%%%%%%%%%%%%%%%% K. Usage of AI Assistants %%%%%%%%%%%%%%%%%%%%
\section{Usage of AI Assistants}
\label{app:ai_assistants}

During the preparation of this work, we utilized AI assistants to refine sentence structure, correct grammatical errors, and improve readability.
We also used them to prepare initial drafts of parts of the appendix, to write scripts for computing the reported statistics, and to translate the Korean examples into English.
The research questions, methodology, and experimental design were developed by the researchers, who reviewed and verified all AI-assisted text, code, and results.
The use of AI assistance did not influence the originality or scientific contributions of the paper.

%% file: Tables/appendix/source_stats.tex
\begin{table}[t]
\centering
\small
\caption{\textbf{Source benchmark statistics.}
Number of questions in each source benchmark and number retained for the knowledge and application analysis.}
\label{tab:source_stats}
\begin{tabular}{@{}lrrrr@{}}
\toprule
 & & \multicolumn{2}{c}{\textbf{KMMLU}} & \\
\cmidrule(lr){3-4}
 & \textbf{CLIcK} & \textbf{Pro} & \textbf{Redux} & \textbf{Total} \\
\midrule
Original questions & 1,995 & 2,822 & 2,587 & 7,404 \\
Excluded           & 62    & 1     & 1     & 64    \\
\midrule
Retained questions & 1,933 & 2,821 & 2,586 & 7,340 \\
\bottomrule
\end{tabular}
\end{table}

%% file: Tables/appendix/ka_taxonomy.tex
\begin{table}[t]
\centering
\small
\caption{\textbf{Knowledge and application taxonomy.}
Definitions of the four levels in each dimension.}
\label{tab:ka_taxonomy}
\begin{tabular}{@{}lp{0.9\columnwidth}@{}}
\toprule
\multicolumn{2}{@{}l}{\textbf{Knowledge Demand (K)}} \\
\midrule
K0 & No external knowledge required \\
K1 & Everyday common sense or general language knowledge \\
K2 & School-level factual or conceptual knowledge \\
K3 & Specialized professional, institutional, or procedural knowledge \\
\midrule
\multicolumn{2}{@{}l}{\textbf{Application Demand (A)}} \\
\midrule
A0 & Direct recall; the retrieved fact or rule is itself the answer \\
A1 & The correspondence between the knowledge and the case is stated explicitly \\
A2 & One piece of knowledge applies; judging whether the case satisfies its condition \\
A3 & Selecting the governing rule among two or more competing candidates \\
\bottomrule
\end{tabular}
\end{table}

%% file: Tables/appendix/k_examples.tex
\begin{table*}[t]
\centering
\small
\caption{\textbf{Examples of knowledge (K) levels.}
One question from CLIcK per level, with our English translation and the reason for its label.
The gold answer is in \textbf{bold}.}
\label{tab:k_examples}
\begin{tabular}{@{}p{0.09\textwidth}p{0.29\textwidth}p{0.31\textwidth}p{0.23\textwidth}@{}}
\toprule
 & \textbf{Question (Korean)} & \textbf{English translation} & \textbf{Why this level} \\
\midrule
K0 \newline No external knowledge
& 제8회 음식 문화 축제. 기간: 2016년 8월 20일(토)--21일(일). 장소: 인주 시청 앞 광장. 행사 내용: 한국 전통요리 무료 시식 및 체험 행사. 체험을 원하시는 분은 현장에서 직접 신청하시기 바랍니다.
\newline 다음 글의 내용과 같은 것을 고르십시오.
\newline A) 축제는 올해 처음 열린다.
\newline \textbf{B) 축제는 주말 이틀 동안 진행된다.}
\newline C) 축제에서 체험을 하려면 예약해야 한다.
\newline D) 축제에서 전통 요리를 사 먹을 수 있다.
& The 8th Food Culture Festival. Dates: Sat., August 20 -- Sun., August 21, 2016. Venue: the plaza in front of Inju City Hall. Events: free tasting of traditional Korean dishes and hands-on activities. Those who wish to take part in the activities should apply on site.
\newline Choose the statement that agrees with the text.
\newline A) The festival is held for the first time this year.
\newline \textbf{B) The festival runs for two days over a weekend.}
\newline C) A reservation is required for the activities.
\newline D) Traditional dishes can be bought and eaten at the festival.
& Everything needed is in the text. \\
\midrule
K1 \newline Everyday knowledge
& 응급환자가 발생했을 때에는 어떤 전화번호로 연락해야 하는가?
\newline A) 112 \quad \textbf{B) 119} \quad C) 118 \quad D) 911
& Which telephone number should be called when there is an emergency patient?
\newline A) 112 \quad \textbf{B) 119} \quad C) 118 \quad D) 911
& Everyday common sense. \\
\midrule
K2 \newline School-level knowledge
& 한국인의 주식인 쌀이 많이 생산되는 평야가 아닌 곳은?
\newline A) 호남평야 \quad B) 나주평야 \newline C) 김포평야 \quad \textbf{D) 강원평야}
& Which of the following is not a plain where rice, the staple food of Koreans, is produced in large quantities?
\newline A) Honam Plain \quad B) Naju Plain \newline C) Gimpo Plain \quad \textbf{D) Gangwon Plain}
& School-level knowledge of geography. \\
\midrule
K3 \newline Specialized knowledge
& 인권 침해 사실이 판단될 경우 국가인권위원회는 어떤 조치를 취하는가?
\newline \textbf{A) 당사자에게 문제를 개선하도록 권고}
\newline B) 법적 처벌
\newline C) 당사자를 해고
\newline D) 사건을 무시
& What measure does the National Human Rights Commission take when it finds that human rights have been violated?
\newline \textbf{A) Recommends that the party remedy the problem}
\newline B) Legal punishment
\newline C) Dismissal of the party
\newline D) Ignoring the case
& Knowledge of the powers of a specific institution. \\
\bottomrule
\end{tabular}
\end{table*}

%% file: Tables/appendix/a_examples.tex
\begin{table*}[t]
\centering
\small
\caption{\textbf{Examples of application (A) levels.}
One question from CLIcK per level, with our English translation and the reason for its label.
The gold answer is in \textbf{bold}.}
\label{tab:a_examples}
\begin{tabular}{@{}p{0.09\textwidth}p{0.29\textwidth}p{0.31\textwidth}p{0.23\textwidth}@{}}
\toprule
 & \textbf{Question (Korean)} & \textbf{English translation} & \textbf{Why this level} \\
\midrule
A0 \newline Knowledge retrieval
& 대한민국 헌법재판소에는 몇 명의 헌법재판관이 있는가?
\newline A) 7명 \quad B) 8명 \quad \textbf{C) 9명} \quad \mbox{D) 10명}
& How many justices sit on the Constitutional Court of Korea?
\newline A) 7 \quad B) 8 \quad \textbf{C) 9} \quad D) 10
& The recalled fact is itself the answer; there is no case to map it onto. \\
\midrule
A1 \newline Explicit mapping
& 이혼 당사자 간의 협의가 잘 이루어지지 않을 때, 자녀의 양육권은 어떻게 결정되는가?
\newline \textbf{A) 가정법원에서 결정한다}
\newline B) 자녀의 의사에 따라 결정한다
\newline C) 변호사가 결정한다
\newline D) 이혼한 당사자 간의 협의에 의해 결정한다
& When the divorcing parties cannot reach an agreement, how is custody of the children decided?
\newline \textbf{A) The family court decides}
\newline B) It is decided by the wishes of the children
\newline C) A lawyer decides
\newline D) It is decided by agreement between the divorced parties
& The question states the condition of the rule (the parties fail to agree), so the rule maps onto the case directly. \\
\midrule
A2 \newline Inferred mapping
& 예금액이 많을 경우 어떻게 해야 모두 보호받을 수 있는가?
\newline A) 한 은행에 모두 예금한다
\newline \textbf{B) 여러 금융 기관에 나누어 예금한다}
\newline C) 금액을 줄인다
\newline D) 보호받을 수 없다
& If one has a large amount of deposits, what should one do so that all of it is protected?
\newline A) Deposit all of it in one bank
\newline \textbf{B) Divide it among several financial institutions}
\newline C) Reduce the amount
\newline D) It cannot be protected
& One rule applies (deposits are protected up to a limit per institution), but it is not stated; one must infer that a large deposit exceeds the limit and what would satisfy the rule. \\
\midrule
A3 \newline Mapping selection
& 다음 정치관과 관련이 깊은 정책으로 옳은 것은?
\newline 임금의 직책은 한 사람의 재상을 논정하는 데 있다 하였으니, 바로 총재를 두고 한 말이다. 총재는 위로는 임금을 받들고 밑으로는 백관을 통솔하여 만민을 다스리는 것이니 직책이 매우 크다. (후략)
\newline A) 육조 직계제의 시행
\newline B) 사간원의 독립
\newline \textbf{C) 의정부 서사제의 시행}
\newline D) 집현전의 설치
& Which policy is closely related to the following view of government?
\newline ``It is said that the duty of the king lies in choosing one chancellor. The chancellor serves the king above and leads all officials below in governing the people, so his office is of great weight.'' (abridged)
\newline A) Direct reporting of the Six Ministries to the king
\newline B) Independence of the Office of the Censor-General
\newline \textbf{C) Deliberation by the State Council}
\newline D) Establishment of the Hall of Worthies
& Two institutions both govern how the king and his ministers share power, with opposite effects: direct reporting strengthens the king, and deliberation by the State Council strengthens the chancellors. One must select which of the two the stated view corresponds to. \\
\bottomrule
\end{tabular}
\end{table*}

%% file: Tables/appendix/ka_reliability.tex
\begin{table}[t]
\centering
\small
\caption{\textbf{Reliability of K/A annotations.}
Linearly weighted Cohen's $\kappa$ on 60 questions, with exact agreement (\%) in parentheses.
LLM--human values are averaged over the two human annotators.}
\label{tab:ka_reliability}
\begin{tabular}{@{}lcc@{}}
\toprule
 & \textbf{K} & \textbf{A} \\
\midrule
Human--Human                & 0.63 (68.3) & 0.60 (56.7) \\
LLM (with silver)--Human    & 0.56 (56.7) & 0.69 (69.2) \\
LLM (without silver)--Human & 0.67 (68.3) & 0.47 (45.0) \\
\bottomrule
\end{tabular}
\end{table}

%% file: Tables/appendix/failure_taxonomy.tex
\begin{table}[t]
\centering
\small
\caption{\textbf{Failure taxonomy.}
Definitions of the 13 failure types and the number of incorrect responses with each primary type ($n=5{,}808$). One response fits none of the types and is counted under other errors.}
\label{tab:failure_taxonomy}
\begin{tabular}{@{}lp{0.68\columnwidth}r@{}}
\toprule
\multicolumn{2}{@{}l}{\textbf{Other errors}} & \textbf{650} \\
\midrule
T0 & Missing or invalid answer output & 262 \\
T1 & Required material (passage, table, or figure) is absent & 45 \\
T2 & Defective question (wrong gold answer, ambiguous or duplicate options, typo) & 342 \\
\midrule
\multicolumn{2}{@{}l}{\textbf{Knowledge failures}} & \textbf{3,422} \\
\midrule
T3 & Recalls a specific fact or value incorrectly & 2,425 \\
T4 & Binds an entity, statute, or event to the wrong identity or attribute & 575 \\
T5 & Required linguistic or structural rule is absent from the reasoning & 27 \\
T6 & Never engages the specific knowledge required; reasoning stays generic & 395 \\
\midrule
\multicolumn{2}{@{}l}{\textbf{Application failures}} & \textbf{1,736} \\
\midrule
T7 & Correct formula and inputs, but incorrect computation & 2 \\
T8 & Correct entities, but incorrect temporal or relational order & 23 \\
T9 & Holds the relevant structural rule but misapplies it & 137 \\
T10 & States the correct rule but misjudges its scope, condition, or exception & 1,401 \\
T11 & Judges each option correctly but ignores the polarity of the question (e.g., ``not'', ``except'') & 170 \\
T12 & Judges each proposition correctly but selects the wrong set in select-all questions & 3 \\
\bottomrule
\end{tabular}
\end{table}

%% file: Tables/appendix/failure_by_a.tex
\begin{table*}[t]
\centering
\small
\caption{\textbf{Failure composition by application level.}
Share of incorrect responses (\%) by primary failure type. Accuracy is averaged over the three models.}
\label{tab:failure_by_a}
\begin{tabular*}{\textwidth}{@{\extracolsep{\fill}}lrrrrrr@{}}
\toprule
 & & & & \multicolumn{3}{c}{\textbf{Failure type (\%)}} \\
\cmidrule{5-7}
\textbf{Level} & \textbf{Questions} & \textbf{Accuracy (\%)} & \textbf{Incorrect responses} & \textbf{Knowledge} & \textbf{Application} & \textbf{Other} \\
\midrule
A0  & 4,482 & 74.5 & 3,433 & 69.8 & 21.1 & 9.1  \\
A1  & 767   & 88.5 & 265   & 41.5 & 36.6 & 21.9 \\
A2  & 884   & 75.1 & 661   & 40.5 & 41.9 & 17.5 \\
A3  & 1,207 & 60.0 & 1,449 & 44.8 & 44.0 & 11.2 \\
\midrule
All & 7,340 & 73.6 & 5,808 & 58.9 & 29.9 & 11.2 \\
\bottomrule
\end{tabular*}
\end{table*}

%% file: Tables/appendix/rubrics.tex
\begin{table*}[t]
\centering
\small
\caption{\textbf{Automated and human validation rubrics.}
A candidate must satisfy every criterion of the LLM rubric (L1--L11) and the human rubric (H1--H5).
Annotators judge H1--H3 before the anchor bundle is revealed and H4--H5 after.}
\label{tab:rubrics}
\begin{tabular}{@{}lll@{}}
\toprule
\textbf{ID} & \textbf{Criterion} & \textbf{Passing condition} \\
\midrule
\multicolumn{3}{@{}l}{\textit{LLM rubric: validity as a question}} \\
\cmidrule{1-3}
L1  & Answer correctness        & The designated gold answer is correct. \\
L2  & Answer uniqueness         & Exactly one option is correct. \\
L3  & Stem determinacy          & The stem admits one reading and supplies everything needed to answer. \\
L4  & Bundle conformance        & No domain knowledge outside the anchor bundle is required. \\
L5  & Surface form              & The text is free of formatting artifacts and orthographic errors. \\
L6  & Target-level structure    & The question has the structure of the target-application level. \\
\cmidrule{1-3}
\multicolumn{3}{@{}l}{\textit{LLM rubric: need for the intended knowledge and application}} \\
\cmidrule{1-3}
L7  & Knowledge preservation    & The question still requires knowledge at the level of the seed. \\
L8  & Shortcut resistance       & The gold answer cannot be isolated without the intended application. \\
L9  & No anchor leakage         & The anchor knowledge is not stated or paraphrased. \\
L10 & Distractor diagnosability & Each distractor results from one identifiable error. \\
L11 & No direct matching        & The question cannot be answered by matching an option to a stated fact. \\
\midrule[\heavyrulewidth]
\multicolumn{3}{@{}l}{\textit{Human rubric: surface quality of the question}} \\
\cmidrule{1-3}
H1  & Korean readability        & The question reads naturally and its meaning is clear. \\
H2  & Natural construction      & The case reads as a realistic situation, not as conditions assembled for the answer. \\
H3  & No surface leakage        & The gold answer cannot be found from length, grammar, or format alone. \\
\cmidrule{1-3}
\multicolumn{3}{@{}l}{\textit{Human rubric: control of knowledge and application}} \\
\cmidrule{1-3}
H4  & Knowledge sufficiency     & The gold answer is supported by the bundle and the case alone. \\
H5  & Target-level structure    & The question has the structure of the target-application level. \\
\bottomrule
\end{tabular}
\vspace{-0.1in}
\end{table*}

%% file: Tables/appendix/construction_config.tex
\begin{table}[t]
\centering
\small
\caption{\textbf{Construction configurations.}
Model and reasoning effort used at each stage of the pipeline.}
\label{tab:construction_config}
\begin{tabular}{@{}lll@{}}
\toprule
\textbf{Stage} & \textbf{Model} & \textbf{Effort} \\
\midrule
Anchor knowledge extraction & GPT-5.5    & high \\
Contrast generation         & GPT-5.5    & high \\
LLM-based auditing          & GPT-5.5    & high \\
Proxy difficulty screening  & GPT-5-nano & high \\
Self-revision               & GPT-5.5    & high \\
Human-guided revision       & GPT-5.5    & high \\
\bottomrule
\end{tabular}
\end{table}

%% file: Tables/appendix/yield_by_type.tex
\begin{table*}[t]
\centering
\small
\caption{\textbf{Yield of the automated stage by transformation type.}
Percentages for the first three candidate rows and the last are shares of the candidates generated; the percentage for self-revision is a share of the candidates revised.}
\label{tab:yield_by_type}
\begin{tabular*}{\textwidth}{@{\extracolsep{\fill}}lrrrr@{}}
\toprule
 & \textbf{A1$\rightarrow$A2} & \textbf{A1$\rightarrow$A3} & \textbf{A2$\rightarrow$A3} & \textbf{All} \\
\midrule
Seed--target pairs                  & 78           & 52           & 165          & 295          \\
Candidates generated                & 531          & 345          & 1,088        & 1,964        \\
\quad All eleven criteria \textsc{pass} & 167 (31.5\%) & 126 (36.5\%) & 334 (30.7\%) & 627 (31.9\%) \\
\quad Proxy answers incorrectly     & 137 (25.8\%) & 125 (36.2\%) & 498 (45.8\%) & 760 (38.7\%) \\
\quad Both (pass before revision)   & 53 (10.0\%)  & 52 (15.1\%)  & 189 (17.4\%) & 294 (15.0\%) \\
\midrule
Candidates revised                  & 477          & 287          & 898          & 1,662        \\
\quad Pass after self-revision      & 82 (17.2\%)  & 62 (21.6\%)  & 185 (20.6\%) & 329 (19.8\%) \\
\midrule
Pass the automated stage            & 135 (25.4\%) & 114 (33.0\%) & 374 (34.4\%) & 623 (31.7\%) \\
Pairs with a passing candidate      & 55           & 44           & 144          & 243          \\
\midrule
Final questions                     & 78           & 51           & 158          & 287          \\
\bottomrule
\end{tabular*}
\end{table*}

%% file: Tables/appendix/human_validation.tex
\begin{table}[t]
\centering
\small
\caption{\textbf{Human validation results by batch.}
Candidates in the first batch are judged by two annotators and those in the second batch by one.}
\label{tab:human_validation}
\begin{tabular}{@{}lrrr@{}}
\toprule
 & \textbf{First} & \textbf{Second} & \textbf{All} \\
\midrule
Candidates validated         & 231 & 70 & 301 \\
\quad Pass directly          & 194 & 58 & 252 \\
\quad Revised                & 37  & 12 & 49  \\
\quad Pass after revision    & 23  & 12 & 35  \\
\midrule
Final questions              & 217 & 70 & 287 \\
\bottomrule
\end{tabular}
\end{table}

%% file: Tables/appendix/criteria.tex
\begin{table}[t]
\centering
\small
\setlength{\tabcolsep}{4pt}
\caption{\textbf{Failures by criterion.}
\textit{Fail} is the share (\%) of candidates that do not pass the criterion: before revision for the LLM rubric (1,964 candidates) and at first validation for the human rubric (301 candidates).
\textit{Resolved} is the share (\%) of these failures that revision fixes, and \textit{New} is the number of failures that revision introduces, with self-revision for the LLM rubric and human-guided revision for the human rubric.}
\label{tab:criteria}
\begin{tabular}{@{}llrrr@{}}
\toprule
\textbf{ID} & \textbf{Criterion} & \textbf{Fail} & \textbf{Resolved} & \textbf{New} \\
\midrule
L1  & Answer correctness        & 1.2  & 96  & 34  \\
L2  & Answer uniqueness         & 1.9  & 100 & 57  \\
L3  & Stem determinacy          & 1.1  & 100 & 37  \\
L4  & Bundle conformance        & 2.2  & 98  & 44  \\
L5  & Surface form              & 0.5  & 89  & 12  \\
L6  & Target-level structure    & 18.5 & 67  & 128 \\
L7  & Knowledge preservation    & 14.5 & 70  & 68  \\
L8  & Shortcut resistance       & 33.5 & 64  & 209 \\
L9  & No anchor leakage         & 27.5 & 78  & 178 \\
L10 & Distractor diagnosability & 17.3 & 90  & 151 \\
L11 & No direct matching        & 6.6  & 92  & 66  \\
\midrule
H1  & Korean readability        & 7.0  & 90  & 1   \\
H2  & Natural construction      & 3.3  & 100 & 2   \\
H3  & No surface leakage        & 5.0  & 87  & 2   \\
H4  & Knowledge sufficiency     & 2.0  & 100 & 0   \\
H5  & Target-level structure    & 4.3  & 77  & 3   \\
\bottomrule
\end{tabular}
\end{table}

%% file: Tables/appendix/anchor.tex
\begin{table}[t]
\centering
\small
\setlength{\tabcolsep}{4pt}
\caption{\textbf{Effect of the anchor bundle.}
Share (\%) of candidates that fail each criterion before revision, when generated with and without the anchor bundle for the same 36 pairs (108 candidates each).
Only L4 differs significantly (Fisher's exact test, $p < 0.001$).}
\label{tab:anchor}
\begin{tabular}{@{}llrr@{}}
\toprule
\textbf{ID} & \textbf{Criterion} & \textbf{With} & \textbf{Without} \\
\midrule
L1  & Answer correctness        & 1.9  & 6.5  \\
L2  & Answer uniqueness         & 3.7  & 6.5  \\
L3  & Stem determinacy          & 1.9  & 3.7  \\
L4  & Bundle conformance        & \textbf{1.9} & \textbf{20.4} \\
L5  & Surface form              & 0.9  & 0.0  \\
L6  & Target-level structure    & 20.4 & 23.1 \\
L7  & Knowledge preservation    & 12.0 & 11.1 \\
L8  & Shortcut resistance       & 25.9 & 30.6 \\
L9  & No anchor leakage         & 22.2 & 24.1 \\
L10 & Distractor diagnosability & 22.2 & 18.5 \\
L11 & No direct matching        & 3.7  & 5.6  \\
\midrule
\multicolumn{2}{@{}l}{Answered correctly by the proxy} & 50.0 & 51.9 \\
\multicolumn{2}{@{}l}{Fail the automated stage}        & 79.6 & 87.0 \\
\bottomrule
\end{tabular}
\end{table}

%% file: Tables/appendix/seed_difficulty.tex
\begin{table}[t]
\centering
\small
\setlength{\tabcolsep}{4pt}
\caption{\textbf{Yield by seed difficulty.}
A1 seeds grouped by the number of Korean LLMs (out of three) that answer the seed incorrectly.
Rates are shares (\%) of the candidates generated.}
\label{tab:seed_difficulty}
\begin{tabular}{@{}lrrrr@{}}
\toprule
\textbf{Incorrect models} & \textbf{0} & \textbf{1} & \textbf{2} & \textbf{3} \\
\midrule
Seeds                                & 12   & 14   & 48   & 24   \\
Seed--target pairs                   & 23   & 28   & 81   & 38   \\
Candidates generated                 & 117  & 246  & 549  & 246  \\
\midrule
All eleven criteria \textsc{pass}    & 58.1 & 43.1 & 31.3 & 28.0 \\
Proxy answers incorrectly            & 12.8 & 30.1 & 28.2 & 39.8 \\
Both (pass before revision)          & 7.7  & 11.8 & 11.8 & 14.2 \\
Pass the automated stage             & 18.8 & 21.5 & 29.9 & 28.9 \\
\bottomrule
\end{tabular}
\end{table}

%% file: Tables/appendix/final_stats.tex
\begin{table}[t]
\centering
\small
\caption{\textbf{Composition of \name{} by application level.}
Number of questions, except for question length, which is the mean number of characters.}
\label{tab:final_stats}
\begin{tabular}{@{}lrrr@{}}
\toprule
 & \textbf{A2} & \textbf{A3} & \textbf{All} \\
\midrule
Questions                 & 78  & 209 & 287 \\
\midrule
\multicolumn{4}{@{}l}{\textit{Source benchmark of the seed}} \\
\quad CLIcK               & 29  & 101 & 130 \\
\quad KMMLU-Pro           & 20  & 70  & 90  \\
\quad KMMLU-Redux         & 29  & 38  & 67  \\
\midrule
\multicolumn{4}{@{}l}{\textit{Knowledge level}} \\
\quad K0                  & 2   & 2   & 4   \\
\quad K1                  & 2   & 27  & 29  \\
\quad K2                  & 31  & 88  & 119 \\
\quad K3                  & 43  & 92  & 135 \\
\midrule
\multicolumn{4}{@{}l}{\textit{Number of options}} \\
\quad Four                & 60  & 122 & 182 \\
\quad Five                & 18  & 87  & 105 \\
\midrule
Question length           & 198 & 349 & 308 \\
\bottomrule
\end{tabular}
\end{table}

%% file: Tables/appendix/eval_models.tex
\begin{table*}[t]
\centering
\small
\setlength{\tabcolsep}{4pt}
\caption{\textbf{Evaluated models and inference settings.}
\textit{Max.} is the limit on output tokens, including reasoning.
Sampling is given as temperature / top-$p$; \textit{default} denotes the default settings of the API.
$^\dagger$Top-$k$ is set to 50.}
\label{tab:eval_models}
\begin{tabular*}{\textwidth}{@{\extracolsep{\fill}}lllllr@{}}
\toprule
\textbf{Model} & \textbf{Identifier} & \textbf{Access} & \textbf{Reasoning} & \textbf{Sampling} & \textbf{Max.} \\
\midrule
K-EXAONE          & \texttt{LGAI-EXAONE/K-EXAONE-236B-A23B} & vLLM       & Enabled & 1.0 / 0.95           & 28,672 \\
A.X K1            & \texttt{skt/A.X-K1}                     & vLLM       & Enabled & 0.6 / 0.95           & 45,056 \\
Solar Open        & \texttt{upstage/Solar-Open-100B}        & vLLM       & High    & 0.8 / 0.95$^\dagger$ & 28,672 \\
\midrule
Aya Expanse 32B   & \texttt{c4ai-aya-expanse-32b}           & Cohere     & None    & 0.3                  & 4,000  \\
\midrule
GLM-4.7           & \texttt{z-ai/glm-4.7}                   & OpenRouter & High    & default              & 32,000 \\
Qwen3.5           & \texttt{qwen/qwen3.5-397b-a17b}         & OpenRouter & High    & default              & 32,000 \\
\midrule
Claude Sonnet 5.5 & \texttt{claude-sonnet-5-5}              & Anthropic  & High    & default              & 32,000 \\
GPT-6.1 Sol       & \texttt{gpt-6.1-sol}                    & OpenAI     & High    & default              & 32,000 \\
\bottomrule
\end{tabular*}
\end{table*}

%% file: Tables/appendix/usage.tex
\begin{table}[t]
\centering
\small
\setlength{\tabcolsep}{4pt}
\caption{\textbf{Token usage of API calls.}
Input and output tokens in millions. Evaluation covers the 287 questions under three conditions.}
\label{tab:usage}
\begin{tabular}{@{}llrrr@{}}
\toprule
\textbf{Stage} & \textbf{Model} & \textbf{Calls} & \textbf{Input} & \textbf{Output} \\
\midrule
\multicolumn{5}{@{}l}{\textit{Construction}} \\
\quad Generation    & GPT-5.5    & 1,965 & 9.3  & 18.9 \\
\quad Auditing      & GPT-5.5    & 3,626 & 17.1 & 27.0 \\
\quad Self-revision & GPT-5.5    & 1,670 & 8.0  & 19.1 \\
\quad Proxy         & GPT-5-nano & 3,626 & 1.3  & 22.5 \\
\midrule
\multicolumn{5}{@{}l}{\textit{Evaluation}} \\
\quad & GPT-6.1 Sol       & 861 & 0.5 & 0.3 \\
\quad & Claude Sonnet 5.5 & 861 & 0.8 & 0.6 \\
\quad & Qwen3.5           & 861 & 0.4 & 3.7 \\
\quad & GLM-4.7           & 861 & 0.6 & 3.4 \\
\quad & Aya Expanse 32B   & 861 & 0.4 & 0.3 \\
\bottomrule
\end{tabular}
\end{table}

%% file: Tables/appendix/by_benchmark.tex
\begin{table*}[t]
\centering
\small
\caption{\textbf{Accuracy (\%) on \name{} by source benchmark of the seed.}
\textit{C} is the \textit{Closed} condition. The numbers of questions are 130 for CLIcK, 90 for KMMLU-Pro, and 67 for KMMLU-Redux.}
\label{tab:by_benchmark}
\begin{tabular*}{\textwidth}{@{\extracolsep{\fill}}lrrrrrrrrrrrr@{}}
\toprule
 & \multicolumn{3}{c}{\textbf{CLIcK}} & \multicolumn{3}{c}{\textbf{KMMLU-Pro}} & \multicolumn{3}{c}{\textbf{KMMLU-Redux}} & \multicolumn{3}{c}{\textbf{All}} \\
\cmidrule{2-4} \cmidrule{5-7} \cmidrule{8-10} \cmidrule{11-13}
\textbf{Model} & \textbf{C} & \textbf{+K} & \textbf{+K+A} & \textbf{C} & \textbf{+K} & \textbf{+K+A} & \textbf{C} & \textbf{+K} & \textbf{+K+A} & \textbf{C} & \textbf{+K} & \textbf{+K+A} \\
\midrule
K-EXAONE          & 50.8 & 76.9 & 86.9 & 40.0 & 78.9 & 90.0 & 47.8 & 88.1 & 95.5  & 46.7 & 80.1 & 89.9 \\
A.X K1            & 47.7 & 66.9 & 77.7 & 31.1 & 71.1 & 83.3 & 47.8 & 74.6 & 92.5  & 42.5 & 70.0 & 82.9 \\
Solar Open        & 35.4 & 65.4 & 74.6 & 23.3 & 67.8 & 83.3 & 35.8 & 74.6 & 88.1  & 31.7 & 68.3 & 80.5 \\
\midrule
Aya Expanse 32B   & 27.7 & 41.5 & 53.8 & 22.2 & 40.0 & 53.3 & 19.4 & 35.8 & 50.7  & 24.0 & 39.7 & 53.0 \\
\midrule
GLM-4.7           & 69.2 & 83.8 & 86.2 & 50.0 & 78.9 & 92.2 & 61.2 & 91.0 & 94.0  & 61.3 & 84.0 & 89.9 \\
Qwen3.5           & 74.6 & 85.4 & 89.2 & 70.0 & 80.0 & 88.9 & 65.7 & 89.6 & 94.0  & 71.1 & 84.7 & 90.2 \\
\midrule
Claude Sonnet 5.5 & 89.2 & 89.2 & 94.6 & 82.2 & 93.3 & 93.3 & 79.1 & 95.5 & 100.0 & 84.7 & 92.0 & 95.5 \\
GPT-6.1 Sol       & 94.6 & 96.2 & 96.9 & 94.4 & 95.6 & 97.8 & 83.6 & 95.5 & 100.0 & 92.0 & 95.8 & 97.9 \\
\bottomrule
\end{tabular*}
\end{table*}

%% file: Tables/appendix/by_family.tex
\begin{table}[t]
\centering
\small
\setlength{\tabcolsep}{4pt}
\caption{\textbf{Accuracy (\%) by model family.}
Averages over the models in each group: Korean (K-EXAONE, A.X K1, Solar Open), multilingual (Aya Expanse 32B), open-weight (GLM-4.7, Qwen3.5), and proprietary (Claude Sonnet 5.5, GPT-6.1 Sol).}
\label{tab:by_family}
\begin{tabular}{@{}lrrrrrr@{}}
\toprule
 & \multicolumn{3}{c}{\textbf{A2}} & \multicolumn{3}{c}{\textbf{A3}} \\
\cmidrule(lr){2-4} \cmidrule(l){5-7}
\textbf{Family} & \textbf{Closed} & \textbf{+K} & \textbf{+K+A} & \textbf{Closed} & \textbf{+K} & \textbf{+K+A} \\
\midrule
Korean       & 49.1 & 86.8 & 92.7 & 37.0 & 67.6 & 81.3 \\
Multilingual & 25.6 & 47.4 & 59.0 & 23.4 & 36.8 & 50.7 \\
Open-weight  & 70.5 & 92.9 & 96.8 & 64.6 & 81.1 & 87.6 \\
Proprietary  & 91.7 & 99.4 & 99.4 & 87.1 & 91.9 & 95.7 \\
\bottomrule
\end{tabular}
\end{table}

%% file: Tables/appendix/knowledge_gain.tex
\begin{table*}[t]
\centering
\small
\caption{\textbf{Effect of adding application statements.}
\textit{Errors} is the number of incorrect responses under \textit{+K}; \textit{Fixed} and \textit{Broken} are the numbers of responses that become correct and incorrect under \textit{+K+A}; $p$ is from the exact McNemar test.}
\label{tab:knowledge_gain}
\begin{tabular*}{\textwidth}{@{\extracolsep{\fill}}lrrrrrrrr@{}}
\toprule
 & \multicolumn{4}{c}{\textbf{A2} ($n=78$)} & \multicolumn{4}{c}{\textbf{A3} ($n=209$)} \\
\cmidrule{2-5} \cmidrule{6-9}
\textbf{Model} & \textbf{Errors} & \textbf{Fixed} & \textbf{Broken} & $p$ & \textbf{Errors} & \textbf{Fixed} & \textbf{Broken} & $p$ \\
\midrule
K-EXAONE          & 6  & 6  & 2 & .289  & 51  & 33 & 9  & $<$.001 \\
A.X K1            & 12 & 8  & 3 & .227  & 74  & 47 & 15 & $<$.001 \\
Solar Open        & 13 & 9  & 4 & .267  & 78  & 43 & 13 & $<$.001 \\
\midrule
Aya Expanse 32B   & 41 & 16 & 7 & .093  & 132 & 57 & 28 & .002 \\
\midrule
GLM-4.7           & 7  & 6  & 2 & .289  & 39  & 18 & 5  & .011 \\
Qwen3.5           & 4  & 3  & 1 & .625  & 40  & 19 & 5  & .007 \\
\midrule
Claude Sonnet 5.5 & 1  & 0  & 0 & 1.000 & 22  & 15 & 5  & .041 \\
GPT-6.1 Sol       & 0  & 0  & 0 & --    & 12  & 8  & 2  & .109 \\
\bottomrule
\end{tabular*}
\end{table*}

%% file: Tables/appendix/seed_results.tex
\begin{table}[t]
\centering
\small
\setlength{\tabcolsep}{4pt}
\caption{\textbf{Accuracy (\%) on the seeds.}
The 165 seeds comprise 58 A1 and 107 A2 questions.}
\label{tab:seed_results}
\begin{tabular}{@{}lrrrrrr@{}}
\toprule
 & \multicolumn{3}{c}{\textbf{A1}} & \multicolumn{3}{c}{\textbf{A2}} \\
\cmidrule(lr){2-4} \cmidrule(l){5-7}
\textbf{Model} & \textbf{Closed} & \textbf{+K} & \textbf{+K+A} & \textbf{Closed} & \textbf{+K} & \textbf{+K+A} \\
\midrule
K-EXAONE   & 39.7 & 84.5 & 87.9 & 32.7 & 73.8 & 83.2 \\
A.X K1     & 43.1 & 84.5 & 87.9 & 30.8 & 68.2 & 76.6 \\
Solar Open & 27.6 & 81.0 & 93.1 & 28.0 & 60.7 & 75.7 \\
\bottomrule
\end{tabular}
\end{table}

%% file: Tables/appendix/failure_contrast.tex
\begin{table}[t]
\centering
\small
\setlength{\tabcolsep}{4pt}
\caption{\textbf{Failure types on \name{}.}
Share (\%) of each primary failure type among the incorrect \textit{Closed} responses; $n$ is the number of incorrect responses.}
\label{tab:failure_contrast}
\begin{tabular}{@{}llrrrr@{}}
\toprule
\textbf{Model} & \textbf{Level} & $n$ & \textbf{Knowledge} & \textbf{Application} & \textbf{Other} \\
\midrule
K-EXAONE   & A2 & 39  & 17.9 & 66.7 & 15.4 \\
           & A3 & 114 & 15.8 & 74.6 & 9.6  \\
A.X K1     & A2 & 34  & 23.5 & 70.6 & 5.9  \\
           & A3 & 131 & 20.6 & 71.8 & 7.6  \\
Solar Open & A2 & 46  & 32.6 & 65.2 & 2.2  \\
           & A3 & 150 & 21.3 & 74.0 & 4.7  \\
\midrule
All        & A2 & 119 & 25.2 & 67.2 & 7.6  \\
           & A3 & 395 & 19.5 & 73.4 & 7.1  \\
\bottomrule
\end{tabular}
\end{table}

%% file: Tables/appendix/matched_test.tex
\begin{table}[t]
\centering
\small
\setlength{\tabcolsep}{4pt}
\caption{\textbf{Matched comparison of A2 and A3 in the \textit{Closed} condition.}
Accuracy (\%) averaged over the 30 seeds with both contrasts, the number of seeds on which each level is more accurate, and the $p$-value of a two-sided sign test.}
\label{tab:matched_test}
\begin{tabular}{@{}lrrrrr@{}}
\toprule
 & \multicolumn{2}{c}{\textbf{Accuracy}} & \multicolumn{2}{c}{\textbf{Seeds favoring}} & \\
\cmidrule(lr){2-3} \cmidrule(lr){4-5}
\textbf{Model} & \textbf{A2} & \textbf{A3} & \textbf{A2} & \textbf{A3} & $p$ \\
\midrule
K-EXAONE          & 46.7 & 40.0 & 9  & 6 & .607 \\
A.X K1            & 51.7 & 21.7 & 15 & 4 & .019 \\
Solar Open        & 38.3 & 23.3 & 8  & 4 & .388 \\
\midrule
Aya Expanse 32B   & 33.3 & 20.0 & 8  & 4 & .388 \\
\midrule
GLM-4.7           & 65.0 & 56.7 & 7  & 3 & .344 \\
Qwen3.5           & 76.7 & 58.3 & 10 & 3 & .092 \\
\midrule
Claude Sonnet 5.5 & 93.3 & 81.7 & 6  & 1 & .125 \\
GPT-6.1 Sol       & 93.3 & 93.3 & 2  & 1 & 1.000 \\
\bottomrule
\end{tabular}
\end{table}

%% file: Tables/appendix/example_contrast.tex
\begin{table*}[t]
\centering
\small
\caption{\textbf{Example of a knowledge-matched contrast.}
An A2 and an A3 contrast from the same seed. The options and the gold answer (A) are the same for both: A) adsorption, B) headspace method, C) solvent extraction, D) steam distillation.}
\label{tab:example_contrast}
\begin{tabular}{@{}p{0.07\textwidth}p{0.43\textwidth}p{0.43\textwidth}@{}}
\toprule
 & \textbf{Korean} & \textbf{English translation} \\
\midrule
Anchor (excerpt)
& 숯(Charcoal)이나 활성탄을 시료 용기의 상부공간에 두어 휘발성 촉진제 성분을 표면에 붙잡아 포집하는 방식은 흡착법이다.
\newline 상부공간법은 시료 위의 기체층인 헤드스페이스 자체를 채취하거나 분석하여 휘발성 성분을 확인하는 방법이다.
& Placing charcoal or activated carbon in the headspace of the sample container, so that volatile accelerant components are captured on its surface, is the adsorption method.
\newline The headspace method identifies volatile components by collecting or analyzing the headspace itself, the gas layer above the sample. \\
\midrule
A2
& 화재 잔류물이 담긴 밀폐병을 실험실로 옮긴 뒤, 분석자는 병마개 안쪽에 작은 검은 탄소성 알갱이가 든 금속망을 매달았다. 병을 따뜻하게 유지하여 휘발 성분이 시료 위 빈 공간으로 퍼지게 한 후, 금속망만 꺼내 분석했더니 휘발성 촉진제 성분이 확인되었다. 이 촉진제 회수 방법으로 가장 적절한 것은?
& After a sealed jar containing fire debris was moved to the laboratory, the analyst suspended a metal mesh holding small black carbonaceous granules inside the lid. The jar was kept warm so that volatile components spread into the empty space above the sample, and when only the mesh was taken out and analyzed, volatile accelerant components were identified. Which method of recovering the accelerant is most appropriate? \\
\midrule
A3
& 화재 잔류물이 든 금속 용기를 온수 욕조에 올려 둔 채 밀봉하였다. 뚜껑 안쪽에는 작은 활성탄 조각을 실로 매달아 잔류물 위 빈 공간에 놓이게 했고, 시간이 지나자 뚜껑 안쪽에는 물방울도 조금 맺혔다. 분석자는 꺼낸 활성탄 조각을 소량의 용매가 든 바이알에 넣은 뒤 그 액을 장비에 주입하였다. 이 절차에서 촉진제 성분을 회수한 방법은 무엇인가?
& A metal container holding fire debris was sealed while placed in a hot water bath. A small piece of activated carbon was hung by a thread inside the lid so that it sat in the empty space above the debris, and after some time a few water droplets also formed inside the lid. The analyst placed the removed piece of activated carbon in a vial containing a small amount of solvent and injected the liquid into the instrument. By which method was the accelerant recovered in this procedure? \\
\bottomrule
\end{tabular}
\end{table*}

%% file: Tables/appendix/example_selfrev.tex
\begin{table*}[t]
\centering
\small
\caption{\textbf{Example of self-revision.}
An A2 candidate that fails shortcut resistance (L8) and its revision. The gold answer is B in both versions.}
\label{tab:example_selfrev}
\begin{tabular}{@{}p{0.07\textwidth}p{0.43\textwidth}p{0.43\textwidth}@{}}
\toprule
 & \textbf{Korean} & \textbf{English translation} \\
\midrule
Before
& Q-system으로 터널 지보량을 검토할 때 ESR을 선정해야 한다. 다음 지하 암반시설들 중 굴착지보비 값을 가장 크게 잡는 것이 타당한 것은?
\newline A) 지하 원자력 발전소
\newline B) 채굴 작업 구간 옆에 마련되어 장비 대기와 버력 적치에 쓰이다가 해당 구간 채굴이 끝나면 되메워질 공동
\newline C) 지하 식품저장 시설
\newline D) 수력발전소의 수로터널
& The ESR must be selected when reviewing tunnel support with the Q-system. For which of the following underground rock facilities is it appropriate to set the largest excavation support ratio?
\newline A) An underground nuclear power plant
\newline B) A cavern prepared beside a mining section, used for equipment standby and muck storage, to be backfilled once mining of the section ends
\newline C) An underground food storage facility
\newline D) The headrace tunnel of a hydropower plant \\
\midrule
Audit (L8)
& \multicolumn{2}{p{0.87\textwidth}@{}}{\textsc{fail}: The gold is isolated by form. B is the only long, detailed situational option, while A, C, and D are short generic facility labels.} \\
\midrule
After
& Q-system으로 터널 지보량을 검토하기 위해 굴착지보비(ESR)를 선정하려고 한다. 다음 지하 암반시설들 중 ESR 값을 가장 크게 잡는 것이 타당한 것은?
\newline A) 지하 원자력 발전소에서 운전 설비를 계속 수용하도록 계획된 주설비 공동
\newline B) 채굴 작업 구간 옆에서 장비와 버력을 두었다가 작업 종료 뒤 되메우도록 계획된 보조 공동
\newline C) 지하 식품저장 시설에서 식품 재고를 계속 수용하도록 계획된 저장 공동
\newline D) 수력발전소에서 발전용 용수를 계속 통수하도록 계획된 수로터널
& The excavation support ratio (ESR) is to be selected in order to review tunnel support with the Q-system. For which of the following underground rock facilities is it appropriate to set the largest ESR?
\newline A) A main cavern in an underground nuclear power plant, planned to keep housing operating equipment
\newline B) An auxiliary cavern beside a mining section, planned to hold equipment and muck and to be backfilled after the work ends
\newline C) A storage cavern in an underground food storage facility, planned to keep housing food stock
\newline D) A headrace tunnel in a hydropower plant, planned to keep conveying water for power generation \\
\bottomrule
\end{tabular}
\end{table*}

%% file: Tables/appendix/example_humanrev.tex
\begin{table*}[t]
\centering
\small
\caption{\textbf{Example of human-guided revision.}
An A2 candidate that fails surface leakage (H3) in human validation and its revision. Place names are fictional. The gold answer is A in both versions.}
\label{tab:example_humanrev}
\begin{tabular}{@{}p{0.07\textwidth}p{0.43\textwidth}p{0.43\textwidth}@{}}
\toprule
 & \textbf{Korean} & \textbf{English translation} \\
\midrule
Before
& 은빛군은 군 명칭과 군청을 그대로 두고 있지만, 행정구역 개편 후 현재 관내 주소가 `청람광역시 은빛군'으로 표시된다. 다만 일부 행정 안내문에는 아직 `백산도 은빛군'이라는 이전 표기가 남아 있다. 은빛군 관내의 통행료를 받지 않는 도로에 설치된 안전표지를 새로 교체하려 한다. 이 안전표지의 도로교통법상 원칙적 설치·관리권자는 누구인가?
\newline A) 청람광역시장이다.
\newline B) 청람광역시 은빛군수이다.
\newline C) 백산도지사이다.
\newline D) 백산도 은빛군수이다.
& Eunbit County keeps its name and county office, but since an administrative reorganization, addresses within it are written ``Eunbit County, Cheongram Metropolitan City.'' Some administrative notices, however, still carry the former notation ``Eunbit County, Baeksan Province.'' A safety sign on a toll-free road in Eunbit County is to be replaced. Under the Road Traffic Act, who is in principle responsible for installing and managing this sign?
\newline A) The mayor of Cheongram Metropolitan City.
\newline B) The head of Eunbit County, Cheongram Metropolitan City.
\newline C) The governor of Baeksan Province.
\newline D) The head of Eunbit County, Baeksan Province. \\
\midrule
Annotator (H3)
& 지문이 현재 주소는 ``청람광역시 은빛군''이고 ``백산도 은빛군''은 ``아직 남아 있는 이전 표기''라고 밝혀, 법령 지식 없이 읽기만으로 `백산도'가 들어간 C와 D가 지워진다.
& The stem states that the current address is ``Eunbit County, Cheongram Metropolitan City'' and that ``Eunbit County, Baeksan Province'' is a former notation that still remains, so C and D, which contain ``Baeksan Province,'' are eliminated by reading alone, without knowledge of the law. \\
\midrule
After
& 은빛군은 군청 명칭과 군수 직함을 계속 사용하지만, 관내 법정 주소는 `청람광역시 은빛군'으로 표시된다. 은빛군 안의 통행료를 받지 않는 도로에 설치된 노후 안전표지를 교체하려 한다. 이 안전표지의 도로교통법상 원칙적 설치·관리권자는 누구인가?
\newline A) 청람광역시장이다.
\newline B) 은빛군수이다.
\newline C) 청람광역시경찰청장이다.
\newline D) 청람광역시 시설관리공단장이다.
& Eunbit County continues to use the name of its county office and the title of county head, but the legal address within it is written ``Eunbit County, Cheongram Metropolitan City.'' An old safety sign on a toll-free road in Eunbit County is to be replaced. Under the Road Traffic Act, who is in principle responsible for installing and managing this sign?
\newline A) The mayor of Cheongram Metropolitan City.
\newline B) The head of Eunbit County.
\newline C) The commissioner of the Cheongram Metropolitan Police Agency.
\newline D) The director of the Cheongram Metropolitan City Facilities Management Corporation. \\
\bottomrule
\end{tabular}
\end{table*}

%% file: Tables/appendix/example_failure.tex
\begin{table*}[t]
\centering
\small
\caption{\textbf{Example of a model failure.}
Responses of A.X K1 to the A3 contrast of Table~\ref{tab:example_contrast} (gold answer: A, adsorption) under the three conditions.}
\label{tab:example_failure}
\begin{tabular}{@{}p{0.11\textwidth}p{0.41\textwidth}p{0.41\textwidth}@{}}
\toprule
 & \textbf{Korean} & \textbf{English translation} \\
\midrule
\textit{Closed}: B \newline (incorrect)
& (모델 추론) 여기서는 용기를 가열하여 증기를 발생시켰고, 그 증기가 뚜껑의 활성탄에 흡착되었다. 따라서 상부공간법을 사용한 것으로 볼 수 있다. 이후 용매로 추출하는 것은 추가적인 과정이지만, 회수 방법 자체는 상부공간법일 것이다.
& (Model reasoning) Here the container was heated to generate vapor, and the vapor was adsorbed onto the activated carbon in the lid. It can therefore be regarded as using the headspace method. The later extraction with a solvent is an additional step, but the recovery method itself would be the headspace method. \\
\midrule
\textit{+K}: C \newline (incorrect)
& (제공된 지식, 발췌) 흡착법은 휘발성 물질이 흡착제의 표면에 달라붙는 성질을 이용하여 성분을 포집·농축하거나 회수하는 방법이다. 상부공간법은 시료 위의 기체층인 헤드스페이스 자체를 채취하거나 분석하여 휘발성 성분을 확인하는 방법이다.
& (Provided knowledge, excerpt) The adsorption method captures, concentrates, or recovers components by using the property of volatile substances to adhere to the surface of an adsorbent. The headspace method identifies volatile components by collecting or analyzing the headspace itself, the gas layer above the sample. \\
\midrule
\textit{+K+A}: A \newline (correct)
& (추가된 적용 진술, 발췌) 용기의 상부공간이 언급되더라도 분류 기준은 위치 자체가 아니라 회수 원리이며, 헤드스페이스를 직접 채취·분석하면 상부공간법이고 숯이나 활성탄 같은 흡착제 표면에 휘발성 성분을 붙잡아 포집하면 흡착법이다.
& (Added application statement, excerpt) Even when the headspace of the container is mentioned, the criterion for classification is the recovery principle rather than the location itself: directly collecting or analyzing the headspace is the headspace method, whereas capturing volatile components on the surface of an adsorbent such as charcoal or activated carbon is the adsorption method. \\
\bottomrule
\end{tabular}
\end{table*}

%% file: Tables/appendix/human_rubric.tex
\begin{table*}[t]
\centering
\small
\caption{\textbf{Instructions for human validation.}
Question and rating levels shown to annotators for each criterion, with our English translation.
The first level of each criterion is the passing level.
H5 is worded differently for A2 and A3 targets.}
\label{tab:human_rubric}
\begin{tabular}{@{}p{0.05\textwidth}p{0.44\textwidth}p{0.44\textwidth}@{}}
\toprule
 & \textbf{Korean} & \textbf{English translation} \\
\midrule
H1
& 한국어 가독성: 문장 자체가 자연스럽고 의미가 명확한가?
\newline \textsc{natural}: 한 번에 자연스럽게 읽힌다.
\newline \textsc{minor\_issue}: 조금 어색하지만 의미는 분명하다.
\newline \textsc{unnatural}: 어색함이나 모호함이 이해를 방해한다.
& Korean readability: Are the sentences themselves natural and is their meaning clear?
\newline \textsc{natural}: Reads naturally at once.
\newline \textsc{minor\_issue}: Slightly awkward, but the meaning is clear.
\newline \textsc{unnatural}: Awkwardness or ambiguity hinders understanding. \\
\midrule
H2
& 출제 인위성: 실제 상황처럼 자연스럽게 읽히는가, 아니면 정답과 선택지를 만들기 위해 조건을 인위적으로 나열한 글처럼 읽히는가?
\newline \textsc{natural}: 현실에서 있을 법한 자연스러운 상황으로 읽힌다.
\newline \textsc{noticeable}: 인위적인 구성 흔적은 있지만 하나의 상황으로 읽힌다.
\newline \textsc{contrived}: 실제 상황보다 문제 풀이를 위한 조건 나열처럼 읽힌다.
& Construction artifice: Does the text read naturally like a real situation, or like a list of conditions arranged to produce the answer and the options?
\newline \textsc{natural}: Reads as a natural situation that could occur in reality.
\newline \textsc{noticeable}: Shows traces of construction but still reads as a single situation.
\newline \textsc{contrived}: Reads as a list of conditions for solving the question rather than a real situation. \\
\midrule
H3
& 표면적 정답 누출: 전문 내용을 몰라도 길이·문법·형식만으로 정답을 찾을 수 있는가?
\newline \textsc{sound}: 표면 단서로 선택지를 줄이기 어렵다.
\newline \textsc{narrowing}: 일부는 지워지지만 둘 이상 남는다.
\newline \textsc{isolating}: 표면 단서만으로 정답 하나가 남는다.
& Surface answer leakage: Can the answer be found from length, grammar, or format alone, without knowing the subject matter?
\newline \textsc{sound}: The options are hard to narrow down with surface cues.
\newline \textsc{narrowing}: Some options are eliminated, but two or more remain.
\newline \textsc{isolating}: Surface cues alone leave only the gold answer. \\
\midrule
H4
& 정답 근거 추적: 정답을 anchor와 사례만으로 뒷받침할 수 있는가?
\newline \textsc{traceable}: 필요한 핵심 근거가 anchor에 있다.
\newline \textsc{partial}: 관련 근거는 있지만 anchor 밖 지식이 더 필요하다.
\newline \textsc{out\_of\_scope}: 핵심 근거를 anchor에서 찾을 수 없다.
& Traceability of the answer: Can the gold answer be supported by the anchor and the case alone?
\newline \textsc{traceable}: The key grounds needed are in the anchor.
\newline \textsc{partial}: Relevant grounds exist, but knowledge outside the anchor is also needed.
\newline \textsc{out\_of\_scope}: The key grounds cannot be found in the anchor. \\
\midrule
H5 (A2)
& 목표 구조: 정확히 하나의 mapping이 적용되고, 조건 충족을 사례에서 추론해야 하는가?
\newline \textsc{clear}: 하나의 mapping이 적용되고 조건 충족을 사례에서 분명히 추론해야 한다.
\newline \textsc{borderline}: 추론은 있으나 조건이 거의 드러나 있거나 mapping 수가 다소 불분명하다.
\newline \textsc{not\_met}: 직접 적용·조회이거나, 여러 mapping이 필요하거나, 사례로 결정할 수 없다.
& Target structure: Does exactly one mapping apply, and must the satisfaction of its condition be inferred from the case?
\newline \textsc{clear}: One mapping applies, and whether its condition is met must clearly be inferred from the case.
\newline \textsc{borderline}: Inference is involved, but the condition is nearly stated or the number of mappings is somewhat unclear.
\newline \textsc{not\_met}: The question is direct application or retrieval, requires several mappings, or cannot be decided from the case. \\
\midrule
H5 (A3)
& 목표 구조: 둘 이상의 mapping이 실제로 경쟁하고, 선택에 따라 정답이 달라지는가?
\newline \textsc{clear}: 둘 이상의 mapping이 사례에 근거하며, 지배 mapping에 따라 정답이 달라진다.
\newline \textsc{borderline}: 두 번째 mapping은 있으나 근거가 약하거나 정답이 달라지는지 불분명하다.
\newline \textsc{not\_met}: 실질적 경쟁이 없거나, mapping이 달라도 정답이 같거나, 결론을 정할 수 없다.
& Target structure: Do two or more mappings genuinely compete, and does the answer change with the choice between them?
\newline \textsc{clear}: Two or more mappings are grounded in the case, and the answer depends on which mapping governs.
\newline \textsc{borderline}: A second mapping exists, but its grounds are weak or it is unclear whether the answer would change.
\newline \textsc{not\_met}: There is no real competition, the answer is the same under either mapping, or the conclusion cannot be determined. \\
\bottomrule
\end{tabular}
\end{table*}